%% file: main.tex
\documentclass[10pt,logo,copyright]{nvidiatechreport}
\input{packages}

\input{common.tex}

\definecolor{darkred}{rgb}{0.7, 0.0, 0.0}

\usepackage{pifont}

\usepackage[nameinlink]{cleveref}
\crefname{equation}{Eq.}{Eqs.}
\crefname{figure}{Fig.}{Figs.}
\crefname{section}{Sec.}{Sec.}
\crefname{appendix}{App.}{App.}
\crefname{table}{Tab.}{Tabs.}
\crefname{algorithm}{Algo}{Algo}
\crefname{thm}{Thm}{Thm}
\Crefname{thm}{Thm}{Thm}
\crefname{prop}{Prop}{Prop}

\newcommand{\crefnames}[3]{%
  \@for\next:=#1\do{%
    \expandafter\crefname\expandafter{\next}{#2}{#3}%
  }%
}

\title{cuVSLAM: CUDA accelerated visual odometry and mapping}

\author[1]{Alexander Korovko}
\author[1]{Dmitry Slepichev}
\author[1]{Alexander Efitorov}
\author[1]{Aigul Dzhumamuratova}
\author[1]{Viktor Kuznetsov}
\author[1]{Joydeep Biswas}
\author[1]{Hesam Rabeti}
\author[1]{Soha Pouya}

\affil[1]{NVIDIA \\
\texttt{\{akorovko, dslepichev, aefitorov, adzhumamurat, vkuznetsov, jbiswas, hrabeti, spouya\}@nvidia.com}}

\begin{abstract}
Accurate and robust pose estimation is a key requirement for any autonomous robot. We present cuVSLAM, a state-of-the-art solution for visual simultaneous localization and mapping, which can operate with a variety of visual-inertial sensor suites, including multiple RGB and depth cameras, and inertial measurement units. cuVSLAM supports operation with as few as one RGB camera to as many as 32 cameras, in arbitrary geometric configurations, thus supporting a wide range of robotic setups. cuVSLAM is specifically optimized using CUDA to deploy in real-time applications with minimal computational overhead on edge-computing devices such as the NVIDIA Jetson. We present the design and implementation of cuVSLAM, example use cases, and empirical results on several state-of-the-art benchmarks demonstrating the best-in-class performance of cuVSLAM.
\end{abstract}

\begin{document}
\maketitle
\abscontent

\section*{Acknowledgements}
We would like to extend our heartfelt acknowledgment to Eugene Vendrovskiy, Stanislav Volodarskiy, and Dmitry Robustov for their invaluable contributions to this project. Their profound expertise, exceptional skills, and commitment to excellence have been pivotal in shaping the outcomes of this work.
\section{Introduction}
Visual Simultaneous Localization and Mapping (VSLAM) is a fundamental capability for autonomous robots, enabling them to build a metric map of their environment while simultaneously determining their location within it. The increasing deployment of autonomous robots across diverse applications, from warehouse automation to last-mile delivery, has created a pressing need for robust, real-time VSLAM solutions that can operate efficiently across diverse settings, while being able to run on energy-efficient edge computing devices.

Traditional VSLAM systems face several key challenges: they often struggle with real-time performance on resource-constrained platforms, have limited flexibility in sensor configurations, and may not fully utilize available hardware acceleration capabilities. Additionally, many existing solutions are optimized for specific sensor configurations, making them less adaptable to diverse robotic platforms with varying sensor requirements.

We evaluate cuVSLAM on standard benchmarks, demonstrating state-of-the-art accuracy while maintaining real-time performance on edge computing platforms. Our results show that cuVSLAM achieves average trajectory errors below 1\% on the KITTI odometry benchmark and position errors under 5cm on the EuRoC dataset, while processing frames in real-time on NVIDIA Jetson platforms. Multi-Stereo mode significantly improves accuracy and robustness on challenging sequences compared to single stereo cameras, as validated through both simulated and real-world dataset evaluations.

We present cuVSLAM, a CUDA-accelerated visual SLAM system designed to address these challenges. cuVSLAM offers several key innovations:

\begin{itemize}
    \item Flexible sensor suite support, accommodating configurations from a single RGB camera to complex arrays of up to 32 cameras with arbitrary geometric arrangements
    \item Support for optional IMU and depth sensors
    \item Efficient CUDA implementation enabling real-time performance on edge devices like NVIDIA Jetson
    \item A modular architecture separating frontend pose estimation from backend map refinement
    \item Robust feature tracking and mapping capabilities that maintain accuracy across diverse environments
\end{itemize}

Our technical approach combines classical SLAM techniques with modern GPU acceleration. The frontend module performs real-time pose estimation using feature detection, tracking, and local mapping, while the backend handles global map consistency through pose graph optimization and loop closure. By leveraging CUDA acceleration throughout the pipeline, from feature detection to bundle adjustment, cuVSLAM achieves superior performance while maintaining high accuracy.

The rest of this technical report is organized as follows: Section 2 describes the system architecture and design, Section 3 presents experimental results and benchmarks, and Section 4 concludes with discussion and future work.

\section{System Architecture and Design}

\begin{figure}[h]
    \centering
	\captionsetup{justification=centering}
    \includegraphics[width=0.75\textwidth]{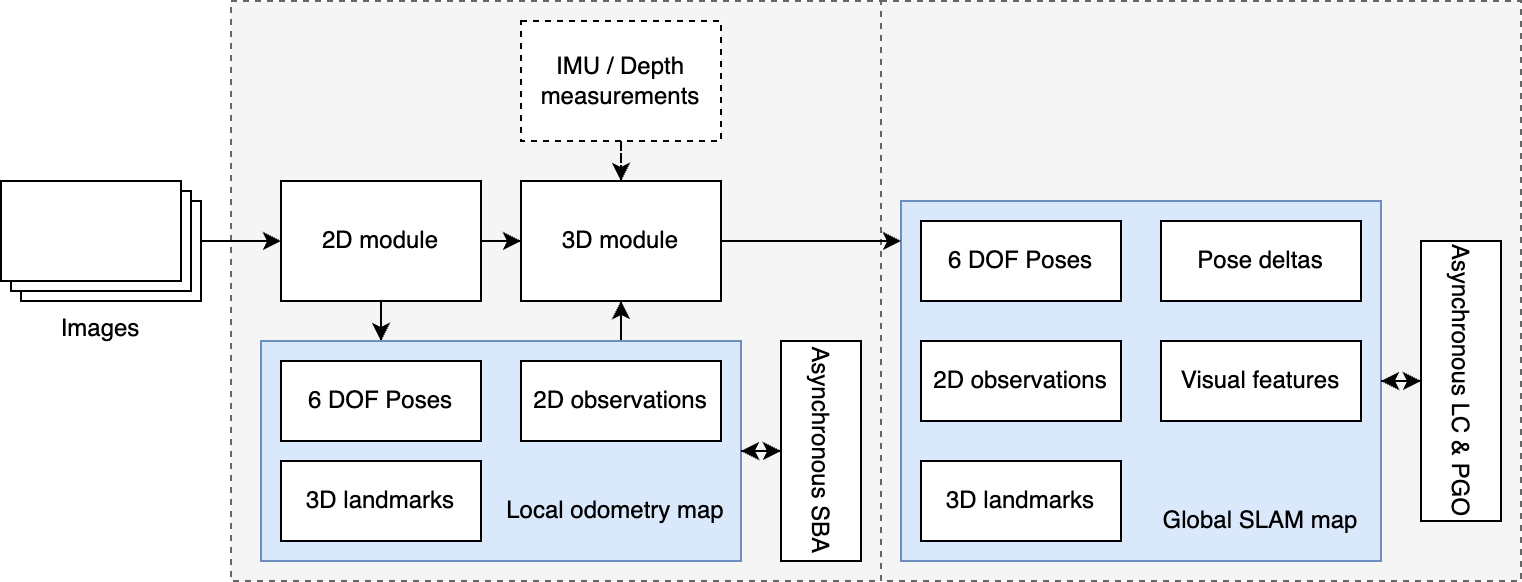}
    \caption{cuVSLAM architecture}
    \label{fig:mesh1}
\end{figure}

The cuVSLAM library implements a general approach to camera pose estimation, where the architecture consists of two major blocks: the frontend, responsible for online pose estimation with minimal latency and maximum throughput, and the backend, focusing on asynchronous and potentially resource-demanding map refinement, loop closing, and pose graph optimization.

The design of the frontend pipeline prioritizes local pose estimation accuracy over global consistency, focusing on delivering smooth trajectories without the bumps introduced by loop closure mechanisms or PGO (Pose Graph Optimization). To achieve this, it maintains a local odometry map consisting of the last $N$ 6DOF camera poses (at keyframes) along with visible 3D landmarks and their observations. Besides the map, the frontend consists of the 2D module, responsible for feature selection, 2D feature tracking, and keyframe selection, and the 3D module, which performs pose estimation based on the local map, 2D information, and other data sources (e.g., IMU or depth measurements).

In contrast to the frontend, cuVSLAM backend focuses on global map and pose consistency. It processes frontend outputs and builds a globally consistent map that integrates camera poses, 2D observations, 3D landmarks, pose deltas, and visual features. The backend maintains global consistency in the map through asynchronous refinement using pose graph optimization and loop closing.
\subsection{2D module} \label{sec:2d_module}

\begin{figure}[h]
    \centering
	\captionsetup{justification=centering}
    \includegraphics[width=0.7\textwidth]{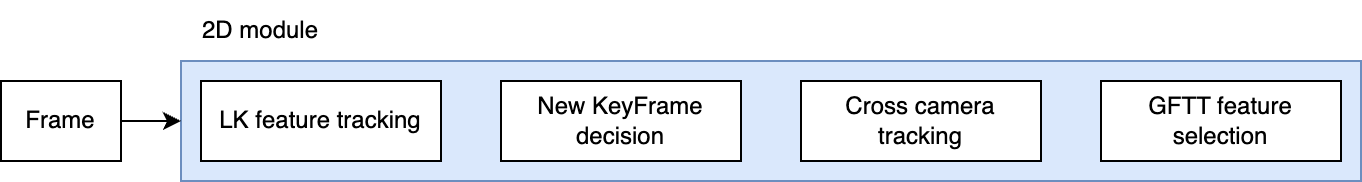}
    \caption{2D module implements keyframe selection, feature selection and 2D tracking.}
    \label{fig:mesh1}
\end{figure}

cuVSLAM implements a set of algorithms operating on the 2D level, which include keypoint selection and feature extraction, keypoint tracking, and keyframe selection.

The pipeline begins with keypoint selection, which is designed to identify high-contrast features and ensure an approximately uniform distribution of keypoints across the image. The image is first divided into non-overlapping patches, forming an $N \times M$ grid. In each patch, the algorithm selects the top $k$ keypoints based on the "Good Features to Track" measure (\cite{GTFF_1993}), where $k$ is calculated such that the total number of keypoints across the entire image exceeds a predefined threshold, $K_I$.

\begin{equation}
k >\left\lfloor \frac{K_I}{N * M} \right\rfloor
\end{equation}

For 2D tracking, cuVSLAM implements a modified version of the Lucas-Kanade algorithm (\cite{LK_1981}, \cite{LK_2000}) for feature tracking that: 1) performs tracking in a coarse-to-fine manner, continuously refining track positions at each image pyramid level, and 2) performs a normalized cross-correlation (NCC) check at each optimization step for each keypoint to filter out unreliable tracks.

If the number of successfully tracked keypoints falls below a threshold, the 2D module creates a new keyframe, which triggers local and global map updates and asynchronous refinement. In stereo and multicamera modes, cuVSLAM launches cross-camera feature tracking to obtain multi-view observations for subsequent landmark triangulation.

\subsection{3D module} \label{sec:3d_module}
When a keyframe is created, the 3D module collects multi-view observations from the 2D module and performs landmark triangulation. The created landmarks, along with their observations, are saved in the local map. After pose estimation, an asynchronous local sparse bundle adjustment (SBA) is initiated for refinement. Given the poses $T_i$, landmarks $p_j$, and 2D observations $o_{ij}$, SBA solves the following optimization task:

\begin{equation}
r^{repr}_{ijk} = \pi(T^{cb}_kT^{bw}_i p^w_j) - o_{j,k}
\end{equation}

\begin{equation}
\hat{T}^{bw}_{1:N},\hat{p}^w_{1:M}  = \arg_{T^{bw}_{1:N},p^w_{1:M}}\min \sum_{i\in [1, N]} \sum_{j\in [1, M]} \sum_{k\in [1, C]} ||\pi(T^{cb}_kT^{bw}_i p^w_j) - o_{j,k}||^2_\Sigma
\end{equation}
where $r^{repr}$ represents reprojection error, $T^{cb}_k$ refers to the k-th camera-from-base transformation, $T^{bw}_i$ denotes the base-from-world transformation, $p^w$ is the 3D landmark in the world frame, and $o_{j,k}$ represents the observation of landmark $j$ by the k-th camera.

Following the common approach, we use the Schur complement to solve for poses $T_i$ first and solve for points $p_j$ afterward. To enhance the efficiency of SBA further, we implement it in CUDA, which is particularly useful for edge devices and/or in multicamera setups. cuVSLAM supports mono, stereo, multicamera, and visual-inertial modes, each of which corresponds to a dedicated 3D solver.
\subsubsection{Stereo} \label{stereo}

Stereo mode is the default mode in cuVSLAM. It utilizes PnP algorithm for pose estimation, that relies on $o_j$, 2D observation, and $p_i$, 3D landmarks, acquired from the local map.
\begin{equation}
\hat{T}^{bw} = \arg_{T^{bw}}\min \sum_{j\in [1, M]} \sum_{k\in [1, 2]} ||r^{repr}_{jk}||^2_\Sigma
\end{equation}

On regular frames, we perform tracking only for the left camera for the tracks that are still active. We query the local map for the corresponding 3D landmarks and use them for pose estimation. When the number of successfully tracked keypoints falls below a threshold, a keyframe is created. After feature selection, we perform cross-camera (left-to-right) tracking to obtain landmark stereo observations and conduct point triangulation. Finally, we add the current camera pose, along with the created landmarks and their observations, into the local map and launch asynchronous SBA for refinement.

\subsubsection{Multi-Stereo}

When a robot operates in environments where it is likely to face a featureless surface from one direction, e.g., narrow corridors or elevators, it becomes vital to use multiple cameras for pose estimation.

cuVSLAM extends the approach introduced in \ref{stereo} to a multicamera setup by treating an arbitrary multiple-camera configuration as a number of stereo pairs.

At startup, we build a Frustum Intersection Graph (FIG)—a directed graph $(V, E)$, where vertices $V$ represent a set of optical sensors, and edges $E$ connect cameras that share a common field of view. Edge direction describes the cross-camera tracking between adjacent cameras, i.e., if $e_{ij}$ exists, cuVSLAM performs 2D tracking from camera $c_i$ to $c_j$.

The FIG is built automatically at cuVSLAM startup. Provided with a list of camera extrinsics/intrinsics, cuVSLAM iterates over each pair of cameras and creates a new edge in the graph when cameras have overlapping fields of view (FoV). To detect this, cuVSLAM places a virtual plane in front of the first camera. It selects uniformly distributed pixels in the image $I_1$, lifts them onto the plane, and backprojects them to the image $I_2$. When the ratio of successfully projected points to the total number of sampled points exceeds a threshold, cuVSLAM detects the shared FoV and creates an edge.

\begin{figure}[h]
    \centering
	\captionsetup{justification=centering}
    \includegraphics[width=0.5\textwidth]{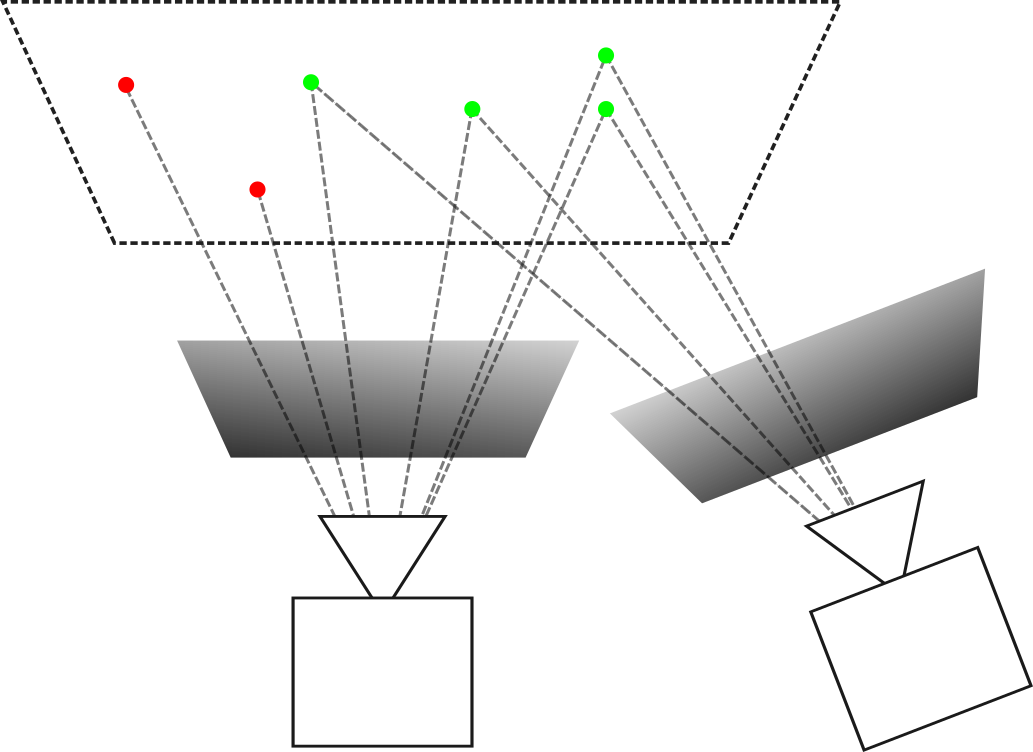}
    \caption{Randomly selected pixels on left picture are lifted on a virtual plane in front of the camera. Pixels that can be successfully projected onto second images are shown in green.}
    \label{fig:mesh1}
\end{figure}

Cameras that correspond to the source of some edge are used for feature tracking/selection over time on each incoming frame. A keyframe is still treated as a global event, while feature selection happens independently for each camera. 

Let $S_{curr}$ be a set of all currently observed points from all the cameras and $S_{kf}$ is the set of points observed in the previous keyframe. Then if:

\begin{equation}
	\frac{\left| S_{curr} \cap S_{kf} \right|}{\left| S_{kf} \right|} < T;
\end{equation}
a keyframe is created and cross-camera tracking is done. Similar to stereo mode we then perform landmark triangulation and refine the map with SBA.

After 2D tracking both on regular frames and keyframes, we collect all the available observations along with the corresponding 3D landmarks for pose estimation.

\begin{equation}
\hat{T}^{bw}  = \arg_{T^{bw}}\min \sum_{j\in [1, M]} \sum_{k\in [1, C]} ||r^{repr}_{jk}||^2_\Sigma
\end{equation}

where $r^{repr}$ denotes a reprojection error for j-th landmark observed from k-th camera in the rig \cite{barfoot_2017}.

Since observations from multiple cameras contribute to pose estimation, cuVSLAM requires images to be synchronized. This is a minor restriction, as many modern cameras provide hardware synchronization capabilities, such as Intel RealSense \cite{RealSense_2018}.

\subsubsection{Visual-Inertial}

\begin{figure}[h]
    \centering
	\captionsetup{justification=centering}
    \includegraphics[width=0.75\textwidth]{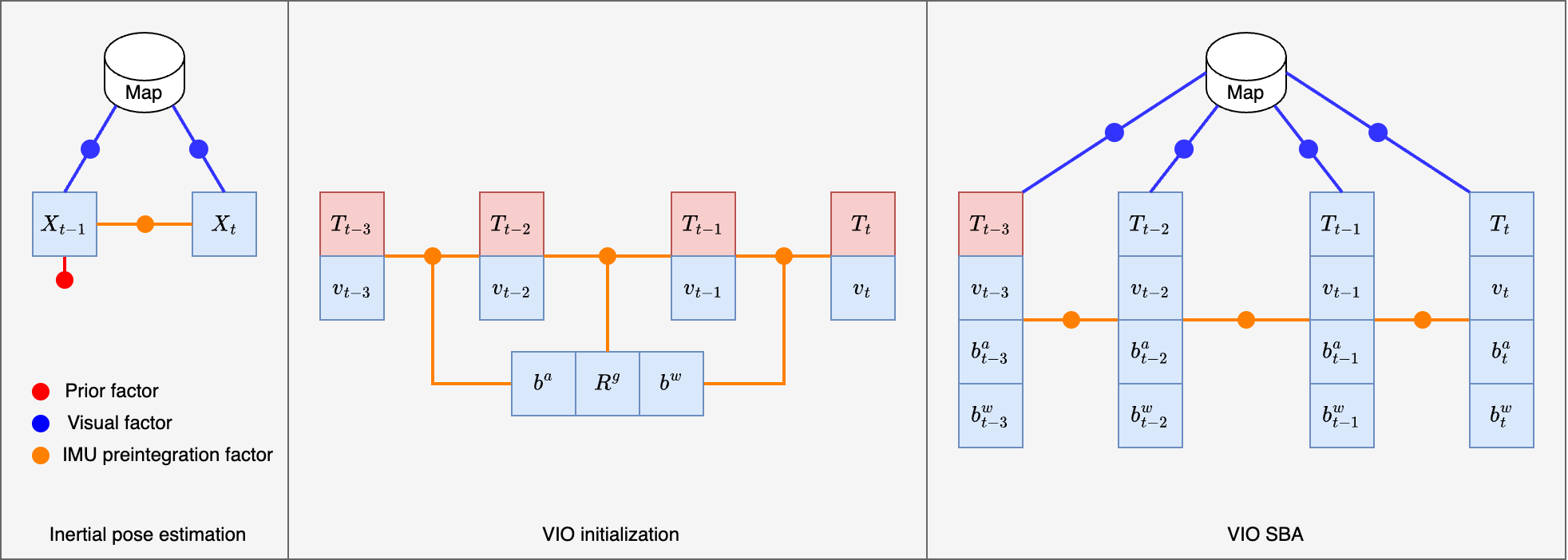}
    \caption{Factor graphs for VIO pipeline.}
    \label{fig:mesh1}
\end{figure}

Visual-Inertial (VI) mode is inspired by \cite{Forster_2017} and \cite{Campos_2021}. Following their approach cuVSLAM defines robot`s state as a combination of 6 DoF pose, linear velocity, accelerometer and gyroscope biases forming a 15 DoF state vector $S = [T \in SE(3), v \in \mathbb{R}^3, b^a \in \mathbb{R}^3, b^w \in \mathbb{R}^3]$.
Robot states along with IMU ($r^{imu}$) and visual ($r^{repr}$) factors are used to build a factor-graph for a list of crucial subproblems, which are:
\begin{itemize}
  \item \textit{Purely visual pose estimation} is used for the initialization of VI pipeline. It consists of the PnP and the asynchronous SBA as it was described in section \ref{stereo} and needed to build a pose history for the subsequent gravity estimation.
  \item \textit{Gravity estimation}. Gravity is a key parameter needed for the IMU preintegration and IMU-aided pose estimation. Assuming the poses predicted by the visual-only part are fixed, the algorithm estimates velocities $v_i$, accelerometer and gyroscope biases $b^a$, $b^w$ and a rotation matrix $R^g$, s.t. $g = R^g [0, 0, 9.81]^T$.
  \begin{equation}
  	R^g, \textbf{b}, v_{0:k} = argmin\left[ \left\| b \right\|^2_{\Sigma_{prior}} + \sum_{i=0}^{k-1} \left\| r^{imu}(S_i, S_{i+1}) \right\|^2_{\Sigma_{IMU}} \right]
  \end{equation}
  \item \textit{Visual-inertial pose estimation} is used every frame. It constrains two consecutive robot states $S_{i-1}$ and $S_{i}$ with IMU and visual factors. In addition $S_{i-1}$ is also constrained by the prior factor.
  \begin{equation}
  	S_{i-1}, S_i = argmin\left[\left\| r^{imu}(S_{i-1}, S_i) \right\|^2_{\Sigma_{IMU}} + \sum_{j=0}^{1}\left\| r^{repr}(S_{i-j}) \right\|^2_{\Sigma_{vis}} + \left\| r^{prior}(S_{i-1}) \right\|^2_{\Sigma_{p}} \right]
  \end{equation}
  \item \textit{Visual-inertial sparse bundle adjustment} is used for local map refinement when IMU data along with gravity and bias estimations are available. Following prior works, we perform relinearization of the IMU factors based on the variable update norm or once per number of iterations.
  \begin{equation}
  	S_{0:k} = argmin\left[ \sum_{i=1}^{k} \left\| r^{imu}(S_{i-1}, S_i) \right\|^2_{\Sigma_{IMU}} + \sum_{i=0}^{k}\left\| r^{repr}(S_i) \right\|^2_{\Sigma_{vis}} \right]
  \end{equation}
\end{itemize}

\subsubsection{Mono-Depth}

\begin{figure}[h]
    \centering
	\captionsetup{justification=centering}
    \includegraphics[width=0.5\textwidth]{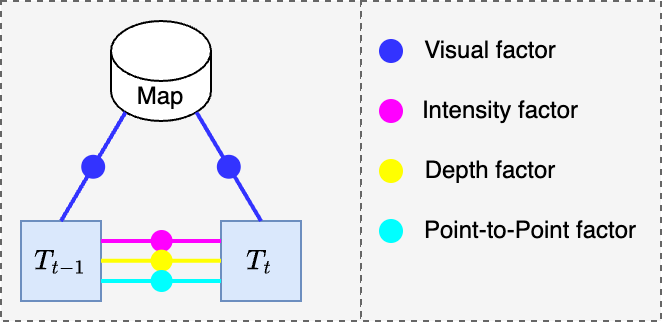}
    \caption{RGBD factor graph.}
    \label{fig:mesh1}
\end{figure}

cuVSLAM supports mono RGBD-based dense frame-to-frame pose estimation. Following  \cite{Dense_RGBD} it considers two consecutive intensity images $I_{1, 2}$ along with pixel-aligned depth images $Z_{1, 2}$ and estimates a relative 6 DoF transformation between corresponding frames.

It defines a list of factors, that constrain frame positions w.r.t each other:
\begin{itemize}
	\item \textbf{Visual factor} used to constrain poses w.r.t the local map by means of reprojection error $r^{repr}$ described earlier.
	\item \textbf{Dense intensity and depth factors} define constraints for each pixel location $x$, introducing the following error functions:
	\begin{equation}
	r^I = \begin{bmatrix}
		I_2(\tau(T_{21}, x)) - I_1 \\
		I_1(\tau(T^{-1}_{21}, x)) - I_2
		\end{bmatrix}
	\end{equation}

	\begin{equation}
		r^Z = \begin{bmatrix}
		Z_2(\tau(T_{21}, x)) - [T_{21}\pi^{-1}(x, Z_1(x))]_z \\
		Z_1(\tau(T^{-1}_{21}, x)) - [T^{-1}_{21}\pi^{-1}(x, Z_2(x))]_z
			\end{bmatrix}
	\end{equation}
	where $T_{21}$ is the relative pose; $\tau(T, x)$ is a warping function, that lifts the pixel location $x$ to 3D, applies transformation $T$ and projects the point back to image plane; $[.]_z$ returns Z coordinate of a point. 
	\item \textbf{Point-to-point factor}. Provided with the 2D-2D matches $(x, y)_{1:N}$ obtained in 2D module, we define the point-to-point error:
 	
 	\begin{equation}
		r^{p} = \pi^{-1}(y, Z_2(y)) - T_{21} \pi^{-1}(x, Z_1(x))
	\end{equation}
\end{itemize}

All together these factors form the combined optimization problem, which we solve using a Levenberg-Marquardt solver implemented on GPU:
\begin{equation}
	\hat{T_{21}} = \arg_{T_{21}} \min (\sum_{p \in Points} ||r^{repr}(p)||^2_\Sigma + \sum_{x \in Pixels} (||r^I(x)||^2 + ||r^Z(x)||_{\sigma_z}^2) + \sum_{(x,y) \in Tracks} ||r^p(x, y)||^2)
\end{equation}

\subsubsection{Mono}

Mono camera mode shares the same common approach as 2D feature tracking, with subsequent point triangulation and pose estimation. A key difference is that triangulation is performed using observations obtained from different camera poses, which we estimate using the same PnP algorithm as described in \ref{stereo}, given the current observations of previously obtained landmarks.

For the first frame pair, when no 3D landmarks are available, cuVSLAM estimates the relative transformation through the fundamental matrix. Specifically, given a set of 2D tracks $(x, y)_{1:N}$, we estimate the fundamental matrix $F$ using RANSAC. With the calibration matrix $K$, we then retrieve the up-to-scale relative pose.

The relative pose is subsequently used to triangulate the first set of landmarks and update the local map, followed by asynchronous SBA refinement.

\subsection{Loop closing and global map refinement}

\begin{figure}[h]
    \centering
	\captionsetup{justification=centering}
    \includegraphics[width=0.75\textwidth]{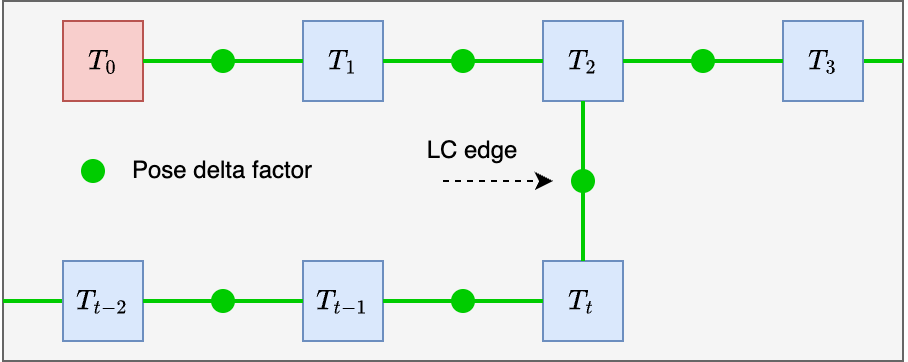}
    \caption{Factor graphs for VIO pipeline.}
    \label{fig:mesh1}
\end{figure}

cuVSLAM loop closing and global map refinement module helps limit global drift by enforcing global consistency of the map. It consists of 6 DoF keyframe poses, landmarks and their observations, pose deltas, and sparse image features. Pose deltas are obtained through the integration of intra-keyframe frontend pose predictions. For image features, we use a list of $9 \times 9$ image patches taken from each level of the image pyramid.
 
We perform the following steps to detect loop closures and refine the map to enforce global consistency:
 \begin{itemize}
 	\item \textbf{Integrate latest data}. When a new keyframe is created, cuVSLAM extracts visual features, integrates intra-keyframe pose predictions, collects currently visible landmarks along with their observations and updates the global map using an asynchronous data queue.
 	\item \textbf{Query nearest neighbors}. After the map update, cuVSLAM performs a pose search of robot poses within a sphere of fixed radius. We use a kd-tree for fast and efficient spatial search.
 	\item \textbf{Landmark selection}. After collecting all the landmarks related to the found poses, we: 1) filter them by geometric visibility, 2) track landmark visual features onto the currently observed image, and 3) filter out landmarks with unsuccessful tracking. Finally, when landmarks from multiple poses are observed, we split the landmarks into sets with respect to connected poses and pick the largest set.
 	\item \textbf{Relative pose estimation}. At this step, we estimate a pose delta $D$ between the pose from the map and the current robot's position. Let $T^{mw}$ denote a pose in the map, connecting the \textit{world} and \textit{map} frames. Then, after shifting landmarks into the \textit{map} frame, we solve for the relative pose estimate:
 	\begin{equation}
 		\hat{T}^{bm} = \arg_{T^{bm}}\min \sum_{j\in [1, M]} \sum_{k\in [1, C]} ||\pi(T^{cb}_kT^{bm} T^{mw} p^w_j) - o_{j,k}||^2_\Sigma
 	\end{equation}
 	
 	\item \textbf{Pose graph optimization}. Provided with the pose deltas between poses either by the frontend or the loop closure mechanism, cuVSLAM refines the map by minimizing the following sum across pose graph edges $E$:
 	\begin{equation}
 		T_{1:N} = \arg_{T_{1:N}} \min \sum_{i, j \in E}||Log(D_{ij}^{-1} T_i^{-1} T_j)||^2
 	\end{equation}
 \end{itemize}

\section{Experiments}
This section demonstrates the capabilities of the cuVSLAM library in robotics applications across diverse real-world environments and camera configurations. We also compare our solution with established classical and deep learning libraries that provide camera pose estimation for long-distance trajectories.

\subsection{System Performance and Hardware Utilization}
We evaluated cuVSLAM performance through two complementary approaches: real-time execution on embedded hardware to measure resource utilization and single-frame processing time, and offline processing of warehouse environment recordings on workstation hardware. The test scenarios involved processing feature-rich images captured from either a wheeled robot navigating a warehouse environment or handheld camera movements. To ensure accurate measurements, all experiments were conducted on systems dedicated solely to cuVSLAM execution, without concurrent resource-intensive operations such as visualization, complex image processing, or data recording that could potentially skew the results.

Table~\ref{tab:performance} summarizes the key performance metrics, with comprehensive results provided in Appendix \ref{sec:hardware_scalability}. The left portion of Table~\ref{tab:performance} presents execution times for cuVSLAM visual tracking function, excluding image acquisition and post-processing overhead. These measurements were obtained using recorded color images at 768×480 resolution (with the exception of mono-depth mode at 640×480 resolution) on two distinct hardware configurations: a desktop workstation featuring an Intel i7-14700 CPU with NVIDIA RTX 4090 GPU, and an embedded Jetson Orin AGX 64GB single-board computer operating in MAXN power mode.

\begin{table}[h]
\centering
\caption{cuVSLAM Performance Evaluation. Left: Per-call processing times measured on RTX 4090 (Desktop) and Jetson AGX Orin (64 GB) using datasets with image resolution 768×480 (except mono-depth mode at 640×480). Right: CPU and GPU utilization on Jetson AGX Orin during live operation with RealSense cameras at 640×480 resolution and 60 FPS (except stereo-inertial mode at 30 FPS).}. 
\label{tab:performance}
\begin{tabular}{lcc|cc}
\hline
\multirow{2}{*}{Configuration} & \multicolumn{2}{c|}{Track call time, ms} & \multicolumn{2}{c}{Jetson HW utilization} \\
\cline{2-5}
 & Desktop & Jetson & CPU \% & GPU \% \\
\hline
Mono & 0.9 & 2.7 & NA & NA \\
Mono-Depth & 5.9 & 15.1 & 2.6 & 55.0 \\
Stereo-Inertial & 1.3 & 3.8 & 1.3 & 2.2 \\
Stereo & 0.4 & 1.8 & 5.5 & 1.7 \\
Multicamera (2-stereo) & 0.8 & 2.0 & 8.3 & 9.0 \\
Multicamera (3-stereo) & 1.4 & 2.3 & 8.3 & 11.0 \\
Multicamera (4-stereo) & 2.1 & NA & NA & NA \\
\hline
\end{tabular}
\end{table}

The right portion of Table~\ref{tab:performance} details CPU and GPU utilization percentages for cuVSLAM visual odometry executed through ISAAC ROS 3.2 with RealSense D435/D455 stereo cameras. All live tests utilized cameras configured at 640×480 resolution and 60 FPS, except for Stereo-Inertial mode, which ran at 30 FPS with IMU data at 200 Hz. The reduced framerate was necessary to ensure valid IMU-based integration, providing 6-7 IMU measurements between each stereo image pair. To accurately isolate cuVSLAM hardware requirements, we employed a controlled experimental methodology: we first established baseline system utilization with the RealSense camera in motion but without cuVSLAM processing, then repeated the similar camera movement pattern with cuVSLAM visual tracking active. Resource utilization metrics were sampled at 15 ms intervals for equal durations in both experiments. The net resource consumption attributable to cuVSLAM was calculated by subtracting baseline measurements from total utilization values and averaging the results. This methodology provides an accurate assessment of the computational overhead introduced by the library. Additional details on evaluation and an exploration of hardware utilization across different resolutions and modes on Jetson devices are provided in Appendix~\ref{sec:hardware_scalability}.

\subsection{Benchmarking}

We evaluated cuVSLAM different operation modes using several publicly available datasets. The Mono-Depth mode was validated on the AR-table~\cite{AR_table_2023}, ICL-NUIM~\cite{Handa_2014} and TUM RGB-D~\cite{Sturm_2012} datasets. The Stereo mode was tested on the EuRoC~\cite{Burri_2016} and KITTI~\cite{Geiger_2013} datasets. The Stereo-Inertial mode was assessed using the EuRoC~\cite{Burri_2016} and TUM-VI~\cite{TUM_VI_2018} datasets. The Multi-Stereo mode was evaluated on simulated TartanAir V2 \cite{TartanAir_v2_2025} and TartanGround~\cite{TartanGround_2025} datasets and our proprietary R2B real-world dataset.

\begin{table}[!h]
\centering
\caption{Evaluation results for cuVSLAM Visual Odometry and SLAM across various operating modes}
\label{tab:benchmarking}
\begin{tabular}{|l|l|l|c|c|c|}
\hline
\textbf{Mode} & \textbf{Dataset} & \textbf{Method} & \textbf{avgRTE, \%} & \textbf{avgRE, deg} & \textbf{RMSE APE} \\
\hline
\multirow{12}{*}{Mono-Depth} & \multirow{4}{*}{AR table} & ORBSLAM3 & - & - & - \\
\cline{3-6}
 & & DPVO & 4.42 & \textbf{1.65} & 0.77 \\
\cline{3-6}
 & & cuVSLAM Odom & 0.34 & 3.59 & 0.09 \\
\cline{3-6}
 & & cuVSLAM SLAM & \textbf{0.19} & 1.68 & \textbf{0.025} \\
\cline{2-6}
& \multirow{4}{*}{ICL-Nuim} & ORBSLAM3 & - & - & - \\
\cline{3-6}
 & & DPVO & 10.28 & \textbf{0.77} & 0.61 \\
\cline{3-6}
 & & cuVSLAM Odom & \textbf{0.41} & 0.99 & \textbf{0.026} \\
\cline{3-6}
 & & cuVSLAM SLAM & 0.44 & 0.97 & \textbf{0.026} \\
\cline{2-6}
 & \multirow{4}{*}{TUM RGB-D} & ORBSLAM3 & \textbf{0.50} & 2.98 & 0.067 \\
\cline{3-6}
 & & DPVO & 13.56 & \textbf{1.98} & 0.80 \\
\cline{3-6}
 & & cuVSLAM Odom & 1.35 & 5.52 & 0.11 \\
\cline{3-6}
 & & cuVSLAM SLAM & 0.99 & 4.13 & \textbf{0.065} \\
\hline
\multirow{8}{*}{Stereo} & \multirow{4}{*}{EuroC*} & ORBSLAM3 & 0.21 & 1.41 & 0.068 \\
\cline{3-6}
 & & DPVO & 0.21 & \textbf{0.96} & 0.10 \\
\cline{3-6}
 & & cuVSLAM Odom & 0.29 & 1.96 & 0.13 \\
\cline{3-6}
 & & cuVSLAM SLAM & \textbf{0.17} & 1.12 & \textbf{0.054} \\
\cline{2-6}
 & \multirow{4}{*}{Kitti} & ORBSLAM3 & 0.31 & 1.20 & 2.98 \\
\cline{3-6}
 & & DPVO & 21.69 & 1.14 & 195.05 \\
\cline{3-6}
 & & cuVSLAM Odom & 0.33 & 1.14 & 3.00 \\
\cline{3-6}
 & & cuVSLAM SLAM & \textbf{0.27} & \textbf{0.93} & \textbf{1.98} \\
\hline
\multirow{8}{*}{Stereo-Inertial} & \multirow{4}{*}{EuroC} & ORBSLAM3 & 5.70 & 72.23 & \textbf{0.066} \\
\cline{3-6}
 & & DPVO & - & - & - \\
\cline{3-6}
 & & cuVSLAM Odom & 0.39 & 2.69 & 0.19 \\
\cline{3-6}
 & & cuVSLAM SLAM & \textbf{0.29} & \textbf{2.27} & 0.13 \\
\cline{2-6}
 & \multirow{4}{*}{TUM-VI Room} & ORBSLAM3 & 2.17 & \textbf{1.37} & \textbf{0.077} \\
\cline{3-6}
 & & DPVO & - & - & - \\
\cline{3-6}
 & & cuVSLAM Odom & 0.20 & 3.85 & 0.18 \\
\cline{3-6}
 & & cuVSLAM SLAM & \textbf{0.12} & 3.00 & 0.12 \\
\hline
\end{tabular}
\end{table}

To quantify visual tracking accuracy, we employed standard metrics: Average Relative Translation (avgRTE) and Rotation (avgRE) Errors~\cite{Geiger_2013}, and Absolute Pose Error (RMSE APE) with trajectory alignment and scale correction~\cite{Sturm_2012} using the EVO library~\cite{Grupp_2017}.

Table~\ref{tab:benchmarking} presents comprehensive benchmarking results across popular datasets from various domains. For comparative analysis, we include performance metrics for the classical computer vision-based ORB-SLAM3 and the deep learning-based DPVO alongside cuVSLAM. Each library was configured according to the dataset characteristics: ORB-SLAM3 and cuVSLAM operated in their corresponding specialized modes (monocular-depth, stereo, or stereo-inertial), while DPVO consistently used monocular mode across all evaluations, utilizing the front-left camera for stereo datasets. Since scale corrections were applied to final trajectories rather than successive pose estimations, translation results may not be meaningful for some datasets; therefore, relative rotation error should be considered the primary metric.

\begin{figure}[!h]
    \centering
	\captionsetup{justification=centering}
    \includegraphics[width=0.75\textwidth]{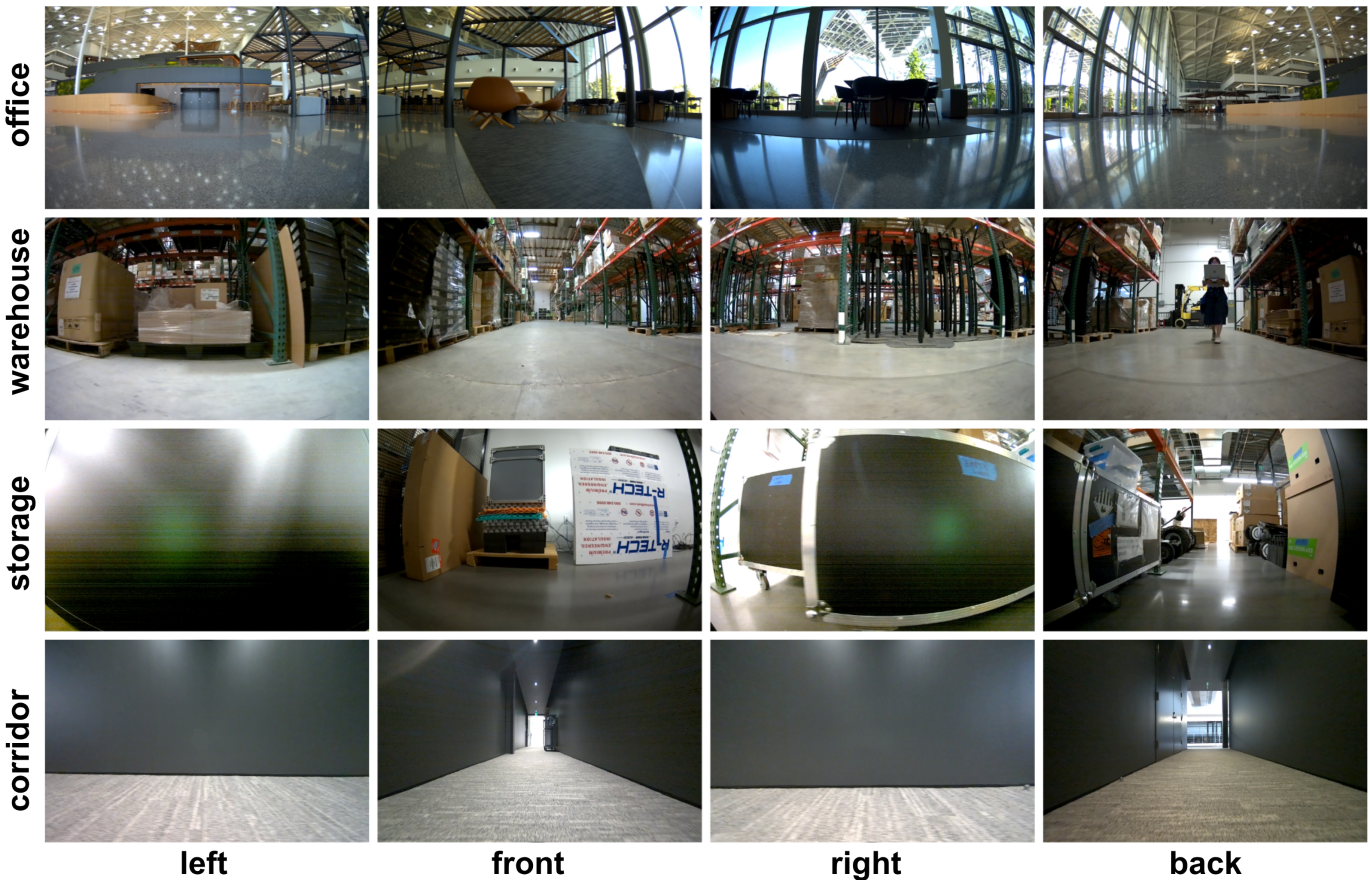}
    \caption{Sample camera views from the R2B dataset captured across diverse environments including large office spaces, industrial warehouses, and narrow corridors in both office and storage areas. The images show simultaneous captures from 4 stereo cameras (displaying only the left image from each stereo pair).}
    \label{fig:r2b_environments}
\end{figure}

To ensure accurate and representative comparisons, we excluded clearly identified outlier and corrupted sequences. For instance, the EuRoC V203 sequence was included only in Stereo-Inertial evaluations, and only a subset of TUM RGB-D sequences were used (see Table~\ref{tab:tum}). For TartanGround datasets, not all sequences provide the full complement of cameras or are available for download. Additionally, many sequences lack overlapping trajectories, preventing loop closure detection—a critical component for evaluating SLAM performance versus odometry. Consequently, we validated SLAM mode on selected subsets of TartanGround (see Table~\ref{tab:tartanairground_slam}) and TartanAir V2 (see Tables~\ref{tab:tartanairv2_odom_1} and~\ref{tab:tartanairv2_odom_2}) datasets.

While the new Tartan datasets provide free camera and quadruped robot movements in diverse simulated environments, rigorous validation of cuVSLAM multi-camera capabilities in real-world scenarios was required. Therefore, a proprietary Multi-Stereo R2B dataset was collected in office and warehouse environments using the Nova Carter wheeled robot equipped with four stereo cameras. The example views from the R2B dataset are shown in \cref{fig:r2b_environments}. This was our only available option for validating this cuVSLAM mode on real data, as publicly available real-world multi-camera datasets either lack hardware-synchronized cameras suitable for stereo pair arrangements~\cite{NCTL_2015,Hilti_2024} or employ software synchronization that cannot ensure simultaneous image capture~\cite{M2dgr_2021,CODa_2023}.

\begin{table}[!h]
    \centering
    \caption{Evaluation results for cuVSLAM Multi-Stereo Visual Odometry and SLAM on multi-camera datasets}
    \label{tab:multistereo_benchmarking}
    \begin{tabular}{|l|l|c|c|c|c|}
    \hline
    \textbf{Dataset} & \textbf{Method} & \textbf{avgRTE, \%} & \textbf{avgRE, deg} & \textbf{RMSE APE} \\
    \hline
    \multirow{2}{*}{R2B} & cuVSLAM Odom & 0.18 & 1.15 & 0.28 \\
    \cline{2-5}
     & cuVSLAM SLAM & \textbf{0.11} & \textbf{0.70} & \textbf{0.18} \\
    \hline
    \multirow{3}{*}{TartanGround} & cuVSLAM Odom* & 0.54 & 0.95 & 0.14 \\
    \cline{2-5}
     & cuVSLAM Odom & 0.21 & 0.48 & 0.09 \\
    \cline{2-5}
     & cuVSLAM SLAM & \textbf{0.17} & \textbf{0.37} & \textbf{0.07} \\
    \hline
    \multirow{3}{*}{TartanAir V2 (Hard)} & cuVSLAM Odom* & 2.56 & 13.97 & 5.20 \\
    \cline{2-5}
     & cuVSLAM Odom & 2.44 & 13.98 & 5.24 \\
    \cline{2-5}
     & cuVSLAM SLAM & \textbf{2.26} & \textbf{12.76} & \textbf{4.99} \\
    \hline
    \end{tabular}
\end{table}

Table~\ref{tab:multistereo_benchmarking} presents cuVSLAM results in Multi-Stereo Mode. Notably, results achieved on the real R2B dataset closely match those from similar domains in the TartanGround dataset, confirming the validity of both datasets and results. For TartanAir V2, we performed detailed validation only on the Hard dataset, which presents greater challenges due to fast six-degrees-of-freedom motion and higher translational and angular velocities. Results for the Easy mode are comparable to those for TartanGround, with a detailed comparison available in Table~\ref{tab:tartanairv2_easy_hard}. While TartanAir V2 results indicate room for improvement, they also demonstrate the importance of multi-camera mode for robust tracking in most environments, with some showing 2--4× improvement in multi-camera SLAM mode.

We anticipate that the recent growth in availability and adoption of synchronized multi-stereo camera systems from various vendors will encourage the creation of new public multi-camera datasets, facilitating further development, testing, and implementation of multi-camera SLAM systems in robotic applications.
\section{Discussion}
We advocate for the wider adoption of multi-camera visual tracking in robotics based on two principal advantages observed in our experiments: enhanced trajectory reliability in feature-poor environments when operating in odometry mode, and increased loop closure detection rates in SLAM mode.

\subsection{Trajectory Reliability Under Visual Occlusion}
\begin{figure}[h]
    \centering
	\captionsetup{justification=centering}
    \includegraphics[width=0.75\textwidth]{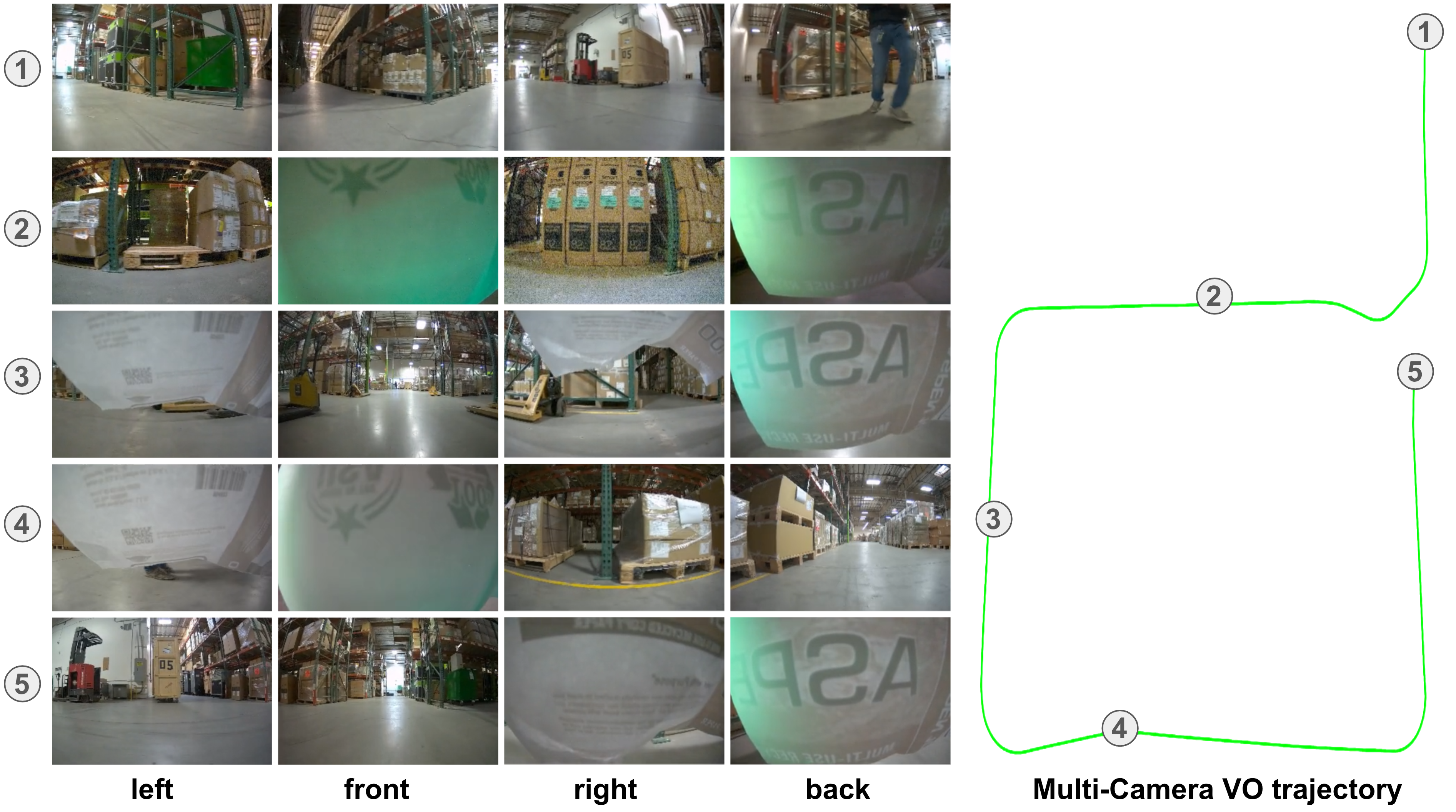}
    \caption{Occlusion resilience test using the Carter robot. cuVSLAM maintains continuous trajectory despite random camera occlusions. Numbered rows show instantaneous views from the robot's 4 stereo cameras (left images only) at corresponding trajectory locations}
    \label{fig:occlusion_test}
\end{figure}

Figure~\ref{fig:occlusion_test} demonstrates the system's trajectory reliability through a practical robustness test. We equipped a Carter robot with 4 stereo cameras and subjected it to challenging visual conditions during continuous motion. Cameras were randomly occluded with opaque film for intervals of 20–60 seconds, with the constraint that at least one stereo pair remained unobstructed at a given moment.

Despite these severe visual constraints, the odometry trajectory maintained stability without exhibiting jumps or discontinuities, validating the system's robustness to partial visual obstruction. This capability represents a significant advancement for visual tracking systems. While single-camera configurations can maintain tracking in feature-poor environments only briefly—typically relying on IMU or depth sensor fusion—our multi-camera approach preserves stable and reliable visual tracking over extended periods even with degraded visual input.

\subsection{Enhanced Loop Closure Detection}
\begin{figure}[!h]
    \centering
	\captionsetup{justification=centering}
    \includegraphics[width=0.75\textwidth]{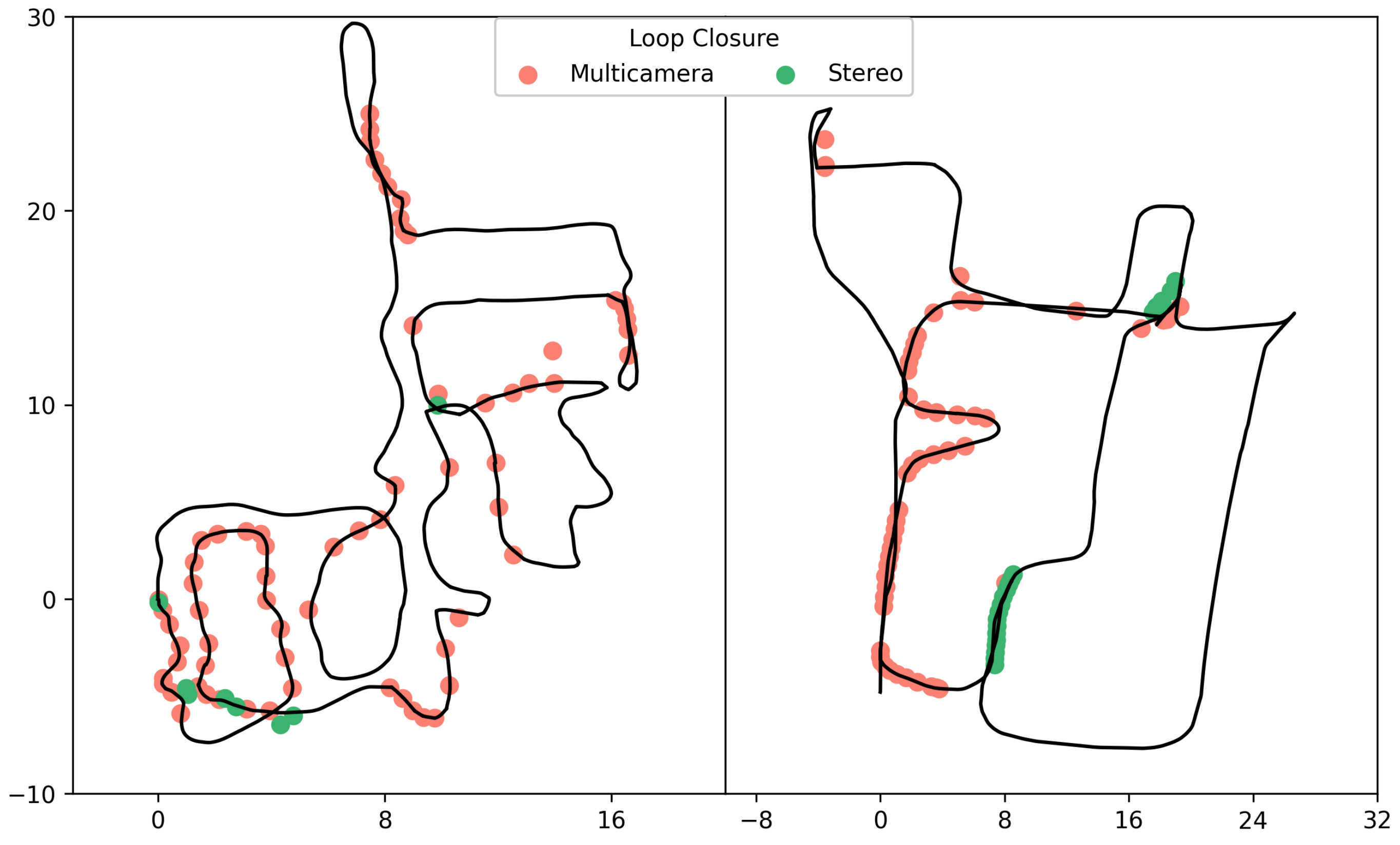}
    \caption{Comparison of loop closure detection events between single-camera and multi-camera (four stereo pairs) SLAM configurations on identical trajectories. The multi-camera setup demonstrates improved mapping consistency due to its wider cumulative field-of-view.}
    \label{fig:loop_closure}
\end{figure}

The expanded collective field of view afforded by multiple cameras directly translates to increased loop closure detection rates. Figure~\ref{fig:loop_closure} presents a comparative analysis of loop closure events between single-stereo and four-stereo camera configurations along identical trajectories. The multi-camera setup demonstrates a substantially higher frequency of successful loop closures, enabling more frequent pose graph optimization.
This increased loop closure rate yields quantitative improvements in localization accuracy. As shown in Table~\ref{tab:multistereo_benchmarking}, evaluation on the Multi-Stereo R2B dataset reveals approximately 40\% improvement in SLAM accuracy metrics compared to pure odometry mode, underscoring the practical benefits of the multi-camera approach for robust localization.

\section{Conclusions}

This technical report demonstrates the robust real-time performance of the cuVSLAM visual tracking library across diverse robotic platforms, including automotive vehicles, aerial drones, wheeled robots, and handheld camera systems. Our comprehensive evaluation encompasses both real-world deployments and synthetic datasets, validating the system's capabilities under varied operational conditions.

The cuVSLAM library offers several key advantages that position it as a production-ready solution for modern robotics applications:

\begin{itemize}
    \item \textbf{Accessibility and Integration}: The library is readily available through both ISAAC ROS\footnote{\url{https://github.com/NVIDIA-ISAAC-ROS/isaac_ros_visual_slam}} and Python\footnote{\url{https://github.com/NVlabs/PyCuVSLAM}} APIs, supporting deployment on both desktop and embedded NVIDIA platforms. This dual-API approach ensures flexibility for developers working across different software ecosystems.

    \item \textbf{Computational Efficiency}: The system maintains low computational overhead while delivering high-accuracy localization, enabling seamless integration into complex robotic stacks such as NVIDIA Isaac Perceptor\footnote{\url{https://developer.nvidia.com/isaac/perceptor}}, which incorporates NVBLOX~\cite{Millane_2023} for 3D reconstruction and Nav2~\cite{Macenski_2020} for autonomous navigation.

    \item \textbf{Minimal Configuration Overhead}: Unlike many SLAM systems that require extensive parameter tuning for different scenarios, cuVSLAM provides robust out-of-the-box performance across diverse environments and sensor configurations, significantly reducing deployment time and complexity.
\end{itemize}

Looking forward, the demonstrated multi-camera capabilities and robust performance characteristics of cuVSLAM establish a strong foundation for next-generation autonomous systems that demand reliable visual localization in challenging real-world environments. The library's proven versatility across platforms and scenarios, combined with its efficient GPU-accelerated implementation, makes it a compelling choice for both research prototypes and production robotic systems.

\clearpage
\setcitestyle{numbers}
\bibliographystyle{plainnat}
\bibliography{main}

\appendix
\section{Appendix}

\subsection{Appendix A. Detailed Description of Dataset Validation}

\subsubsection{KITTI Dataset}
\begin{figure}[!h]
    \centering
	\captionsetup{justification=centering}
    \includegraphics[width=0.95\textwidth]{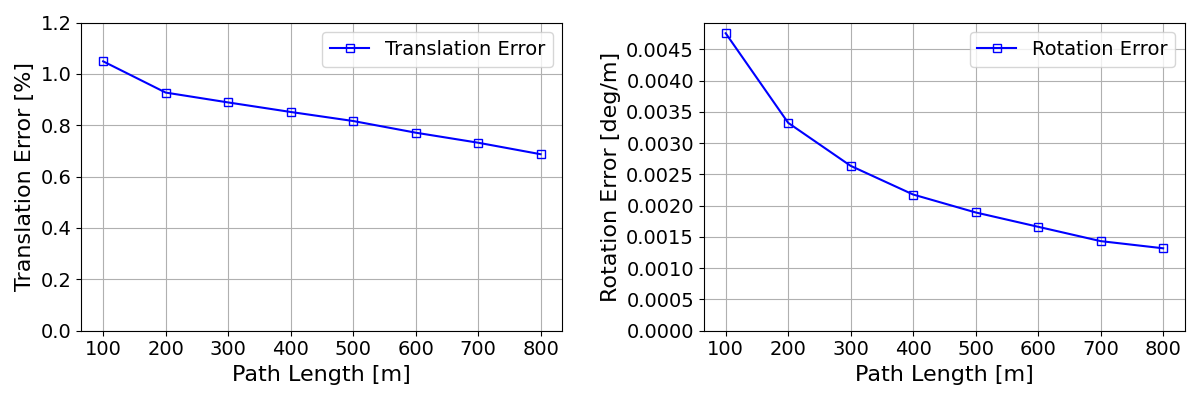}
    \caption{cuVSLAM evaluation results on KITTI odometry benchmark sequences 00-10. Translation and rotation errors are calculated on segments of 100-800m following the KITTI
public leaderboard methodology, averaged metrics: \textbf{Translation: 0.85\%, Rotation: 0.0025 [deg/m]}.}
    \label{fig:kitti_benchmark}
\end{figure}

We evaluated our method on sequences 0-10 of the KITTI dataset, for which ground truth trajectories are available. For comparison with the conventional KITTI leaderboard~\cite{KITTI_Odometry_2012}, Figure~\ref{fig:kitti_benchmark} presents relative translation (avgRTE) and rotation (avgRE) errors calculated on trajectory segments of lengths 100, 200, 300, 400, 500, 600, 700, and 800 meters. In contrast Table~\ref{tab:benchmarking} presents results calculated without trajectory segmentation to maintain metric comparability across different datasets. 

\subsubsection{TUM-VI Room Dataset}
To validate the stereo-inertial tracking mode on real hardware under aggressive camera motions, we evaluated cuVSLAM on the Room sequences of the TUM-VI dataset~\cite{TUM_VI_2018}. These sequences were selected because they provide complete ground truth trajectories. We used undistorted images with a resolution of 512x512 pixels and applied a 50-pixel mask along each frame edge to mitigate significant fisheye lens distortion, which negatively affects tracking performance. To facilitate comparison with other visual tracking methods, we followed the evaluation protocol described in the original TUM-VI paper~\cite{TUM_VI_2018}, computing the average RMSE of the Relative Pose Error (RPE) in meters/degrees over 1-second segments. Detailed results are provided in Table~\ref{tab:tum_vi_room}.
\begin{table}[!h]
    \centering
    \caption{cuVSLAM Stereo Inertial tracking validation results on TUM-VI Room dataset RMSE RPE (m / deg) on 1 second segments}
    \label{tab:tum_vi_room}
    \begin{tabular}{|l|c|c|}
    \hline
    \textbf{Sequence} & \textbf{ODOM} & \textbf{SLAM} \\
    \hline
    room1 & 0.023 / 0.89 & 0.025 / 0.90 \\
    \hline
    room2 & 0.048 / 2.28 & 0.048 / 2.29 \\
    \hline
    room3 & 0.047 / 1.93 & 0.048 / 1.95 \\
    \hline
    room4 & 0.015 / 0.54 & 0.015 / 0.55 \\
    \hline
    room5 & 0.019 / 0.56 & 0.021 / 0.57 \\
    \hline
    room6 & 0.021 / 0.57 & 0.021 / 0.58 \\
    \hline
    \end{tabular}
    \end{table}

\subsubsection{AR-table Dataset}

This dataset contains sequences captured using an Intel RealSense D455 depth camera, providing image and depth data at a resolution of 848x480 pixels. The enhanced image and depth quality allows achieving performance on this real-world camera close to that obtained on the ideal simulated ICL-NUIM dataset, as shown in Table~\ref{tab:benchmarking}.

However, consistent gaps in depth values were observed along the sides and bottom edges of the depth images. To mitigate this issue, we applied static masking of 24 pixels on each side and 10 pixels at the bottom. Additionally, several prolonged depth-image dropouts (lasting more than one second) occurred in sequences 3, 4, 5, and 8. To address these interruptions, we segmented the affected sequences into continuous portions, executed tracking independently on each segment, and subsequently merged the results.

To facilitate comparison with other visual tracking methods, we followed the evaluation methodology described in the original AR-table dataset paper~\cite{AR_table_2023}. Specifically, we computed Absolute Trajectory Error (ATE), reporting translation errors in centimeters and rotation errors in degrees. The obtained results are summarized in Table~\ref{tab:ar_table}.
\begin{table}[!h]
    \centering
    \caption{cuVSLAM Mono-Depth Odometry and SLAM validation results on AR table dataset ATE (degree / cm)}
    \label{tab:ar_table}
    \scriptsize
    \resizebox{\textwidth}{!}{%
    \begin{tabular}{|l|c|c|c|c|c|c|c|c|}
    \hline
    \textbf{Mode} & \textbf{table\_01} & \textbf{table\_02} & \textbf{table\_03} & \textbf{table\_04} & \textbf{table\_05} & \textbf{table\_06} & \textbf{table\_07} & \textbf{table\_08} \\
    \hline
    ODOM & 3.77 / 2.03 & 9.36 / 5.41 & 6.09 / 2.28 & 3.13 / 2.21 & 2.27 / 1.97 & 3.43 / 3.39 & 18.5 / 6.17 & 13.2 / 8.18 \\
    \hline
    SLAM & 2.98 / 1.76 & 2.23 / 1.77 & 2.11 / 0.87 & 2.12 / 1.64 & 1.24 / 1.22 & 1.93 / 1.23 & 8.88 / 3.82 & 4.83 / 3.69 \\
    \hline
    \end{tabular}%
    }
\end{table}

\subsubsection{TUM RGB-D Dataset}
To validate the Mono-Depth mode on this dataset, a subset of 10 sequences from the Freiburg3 collection was used. These sequences were selected as they provide rectified images, which are required by cuVSLAM Mono-Depth mode. Minimal preprocessing was performed, consisting only of depth-image pair synchronization without further image or depth processing. The complete list of evaluated sequences is provided in Table~\ref{tab:tum}.

\begin{table}[!h]
    \centering
    \caption{Results of cuVSLAM Mono-Depth Odometry and SLAM evaluation on TUM RGB-D dataset}
    \label{tab:tum}
    \begin{tabular}{|l|l|c|c|c|}
    \hline
    \textbf{Sequence} & \textbf{Mode} & \textbf{avgRTE} & \textbf{avgRE} & \textbf{RMSE APE} \\
    \hline
    \multirow{2}{*}{large cabinet validation} & ODOM & 1.74 & 2.08 & 0.12 \\
     & SLAM & 1.53 & 1.60 & 0.11 \\
    \hline
    \multirow{2}{*}{long office household} & ODOM & 1.27 & 7.89 & 0.20 \\
     & SLAM & 0.96 & 5.94 & 0.06 \\
    \hline
    \multirow{2}{*}{nostructure texture far} & ODOM & 1.29 & 1.66 & 0.07 \\
     & SLAM & 1.44 & 1.63 & 0.06 \\
    \hline
    \multirow{2}{*}{nostructure texture near withloop} & ODOM & 1.23 & 4.62 & 0.10 \\
     & SLAM & 1.25 & 4.71 & 0.04 \\
    \hline
    \multirow{2}{*}{sitting halfsphere} & ODOM & 2.07 & 3.67 & 0.12 \\
     & SLAM & 0.61 & 1.42 & 0.05 \\
    \hline
    \multirow{2}{*}{sitting xyz} & ODOM & 1.03 & 1.21 & 0.05 \\
     & SLAM & 0.65 & 0.86 & 0.03 \\
    \hline
    \multirow{2}{*}{sitting xyz validation} & ODOM & 0.90 & 1.42 & 0.05 \\
     & SLAM & 0.44 & 0.67 & 0.03 \\
    \hline
    \multirow{2}{*}{structure texture far} & ODOM & 1.11 & 1.46 & 0.04 \\
     & SLAM & 0.55 & 0.69 & 0.02 \\
    \hline
    \multirow{2}{*}{structure texture near} & ODOM & 1.45 & 2.15 & 0.03 \\
     & SLAM & 1.45 & 2.09 & 0.03 \\
    \hline
    \multirow{2}{*}{teddy} & ODOM & 1.38 & 29.04 & 0.31 \\
     & SLAM & 1.04 & 21.70 & 0.22 \\
    \hline
    \end{tabular}
\end{table}

\subsubsection{TartanGround and TartanAir V2} \label{sec:tartanairground_description}
\textbf{Odometry Mode.}
We validated cuVSLAM using 216 sequences from TartanGround ~\cite{TartanGround_2025} and 539 sequences from TartanAir V2 ~\cite{TartanAir_v2_2025} Hard datasets. Complete sequence lists are provided in Tables \ref{tab:tartanairground_odom} and \ref{tab:tartanairv2_odom_1}, respectively. All results were obtained using a four stereo-camera configuration (front, back, left, and right) with 640×640 undistorted images.
While the TartanAir V2 dataset supports two additional cameras (top and bottom), their inclusion yielded minimal metric improvement (less than 5\%). For the TartanGround dataset, we observed slight performance degradation when using these additional cameras, attributed to meaningless feature selection from sky regions and repetitive floor textures in planar robot navigation areas.

To facilitate comparison with other visual tracking libraries, we computed additional metrics—relative translation ($t_{\mathrm{rel}}$) and rotation ($r_{\mathrm{rel}}$) per frame—using scripts from the official TartanAir GitHub repository~\cite{TartanGroundRepo_2025}. Since not all environments from the original TartanGround paper were available, we present environment-averaged results in Table~\ref{tab:tartanairground_odom} to enable future comparative studies.
\begin{table}[h]
    \centering
    \caption{Odometry evaluation results for cuVSLAM Multi-Stereo configuration (4 stereo cameras) on TartanGround dataset. Values represent averages computed per environment.}
    \label{tab:tartanairground_odom}
    \begin{tabular}{|l|l|c|c|c|c|c|}
    \hline
    Environment & Sequences & avgRTE & avgRE & RMSE APE & $t_{\mathrm{rel}}$ & $r_{\mathrm{rel}}$ \\
    \hline
    Downtown & P0-P17 & 0.30 & 0.41 & 0.11 & 0.0019 & 0.00013 \\
    ForestEnv & P0-P22 & 0.45 & 0.72 & 0.19 & 0.0024 & 0.00036 \\
    Gascola & P0-P21 & 0.24 & 0.46 & 0.09 & 0.0012 & 0.00018 \\
    GreatMarsh & P1-P19, P21-P29 & 2.34 & 4.21 & 0.45 & 0.0037 & 0.00126 \\
    ModernCityDowntown & P1-P10 & 0.19 & 0.39 & 0.08 & 0.0012 & 0.00014 \\
    ModularNeighborhood & P0-P29 & 0.42 & 0.65 & 0.12 & 0.0014 & 0.00016 \\
    NordicHarbor & P0-P27 & 0.23 & 0.41 & 0.06 & 0.0010 & 0.00015 \\
    OldTownFall & P0-P7 & 0.15 & 0.34 & 0.04 & 0.0011 & 0.00013 \\
    OldTownSummer & P0-P7 & 0.15 & 0.34 & 0.05 & 0.0011 & 0.00013 \\
    SeasonalForestAutumn & P0-P19 & 0.27 & 0.42 & 0.11 & 0.0017 & 0.00022 \\
    SeasonalForestSpring & P0-P19 & 0.23 & 0.49 & 0.12 & 0.0016 & 0.00024 \\
    SeasonalForestWinter & P0-P21 & 0.27 & 0.51 & 0.08 & 0.0011 & 0.00018 \\
    \hline
    \textbf{Average} &  & \textbf{0.54} & \textbf{0.95} & \textbf{0.14} & \textbf{0.0017} & \textbf{0.00032} \\
    \hline
    \end{tabular}
    \end{table}

\begin{figure}[h]
    \centering
	\captionsetup{justification=centering}
    \includegraphics[width=1.0\textwidth]{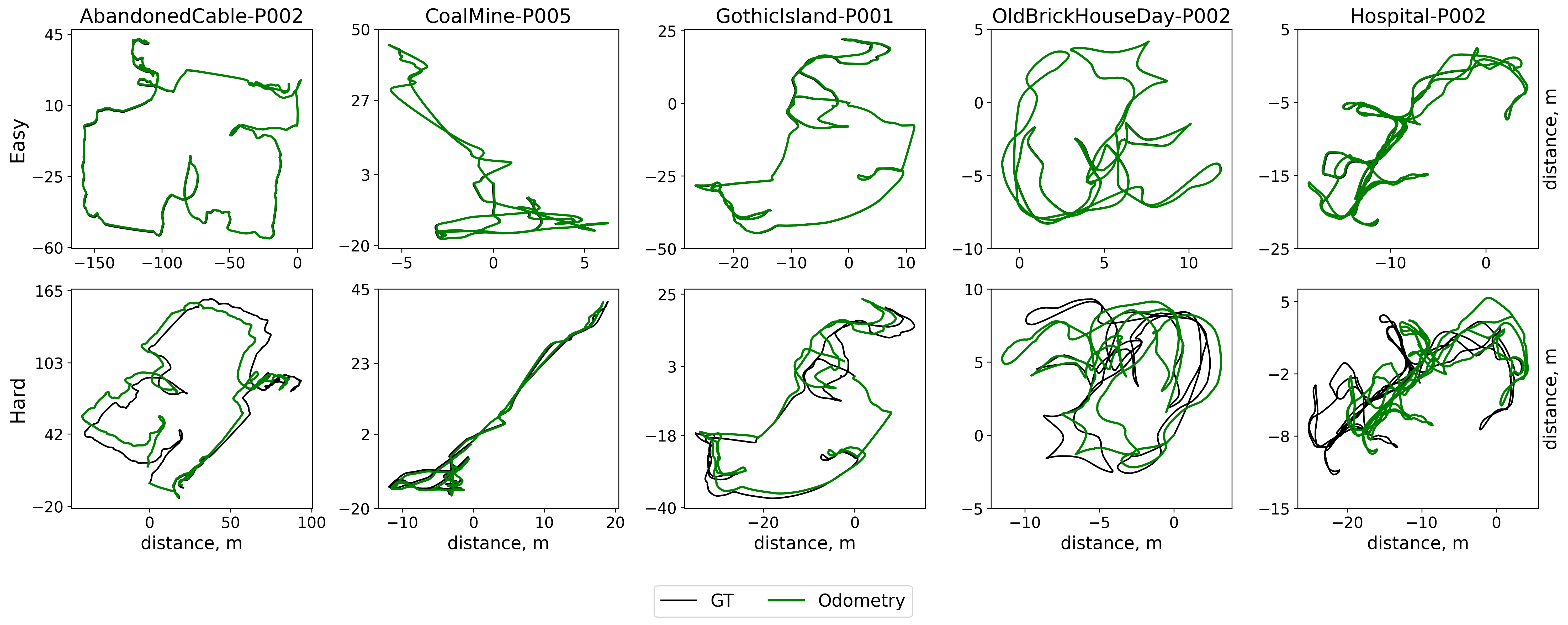}
    \caption{Example trajectories predicted by cuVSLAM (4 Multi-Stereo Odometry) on TartanAir V2 (Easy and Hard). Black lines represent ground truth, green lines represent odometry predictions. Columns correspond to the same environment and sequence label at different difficulties.}
    \label{fig:tartanV2_easy_hard_trajs}
\end{figure}

TartanAir V2 Easy sequences were excluded from our analysis as they present minimal challenge for cuVSLAM, yielding consistently high tracking quality similar to the TartanGround dataset. To support this decision, Table~\ref{tab:tartanairv2_easy_hard} presents a comparison of averaged four-stereo-camera odometry results between Easy and Hard subsets of TartanAir V2, with example trajectory estimates shown in Figure~\ref{fig:tartanV2_easy_hard_trajs}.

\begin{table}[h]
    \centering
    \caption{Comparative odometry evaluation of cuVSLAM Multi-Stereo (4 stereo cameras) on TartanAir V2 Easy and Hard sequences. Results show average performance metrics for selected environments.}
    \label{tab:tartanairv2_easy_hard}
    \begin{tabular}{|l|c|c|c|c|}
    \hline
    \textbf{Environment} & \textbf{Difficulty} & \textbf{avgRTE} & \textbf{avgRE} & \textbf{RMSE APE} \\
    \hline
    \multirow{2}{*}{AbandonedCable} & easy & 0.05 & 0.56 & 0.26 \\
     & hard & 1.93 & 14.5 & 8.65 \\
    \hline
    \multirow{2}{*}{AbandonedFactory2} & easy & 0.02 & 0.13 & 0.02 \\
     & hard & 0.27 & 1.41 & 0.34 \\
    \hline
    \multirow{2}{*}{CoalMine} & easy & 0.03 & 0.16 & 0.02 \\
     & hard & 0.37 & 1.25 & 0.56 \\
    \hline
    \multirow{2}{*}{GothicIsland} & easy & 0.08 & 0.58 & 0.17 \\
     & hard & 1.73 & 13.44 & 4.48 \\
    \hline
    \multirow{2}{*}{Hospital} & easy & 0.04 & 0.60 & 0.06 \\
     & hard & 1.50 & 16.17 & 3.25 \\
    \hline
    \multirow{2}{*}{JapaneseAlley} & easy & 0.06 & 0.34 & 0.06 \\
     & hard & 0.97 & 5.45 & 0.50 \\
    \hline
    \multirow{2}{*}{OldBrickHouseDay} & easy & 0.06 & 0.54 & 0.04 \\
     & hard & 1.74 & 11.15 & 1.48 \\
    \hline
    \end{tabular}
    \end{table}

\textbf{SLAM Mode.}
To evaluate the impact of SLAM mode on trajectory estimation, we analyzed only sequences containing loop closures. This resulted in a subset of 32 sequences for TartanGround and 506 sequences for TartanAir V2 Hard. Complete tracking results are provided in Tables~\ref{tab:tartanairground_slam} and~\ref{tab:tartanairv2_odom_1}, with all results averaged across environments.    

For the TartanAir V2 dataset, most environments utilized all available sequences. Sequences excluded from SLAM validation are marked with an asterisk (P*), while their corresponding visual odometry results are labeled ODOM*, indicating performance metrics computed across all available sequences for those specific environments, including the excluded sequences.

\subsubsection{Mutli-Stereo R2B Dataset}
    
\begin{figure}[h]
    \centering
	\captionsetup{justification=centering}
    \includegraphics[width=0.95\textwidth]{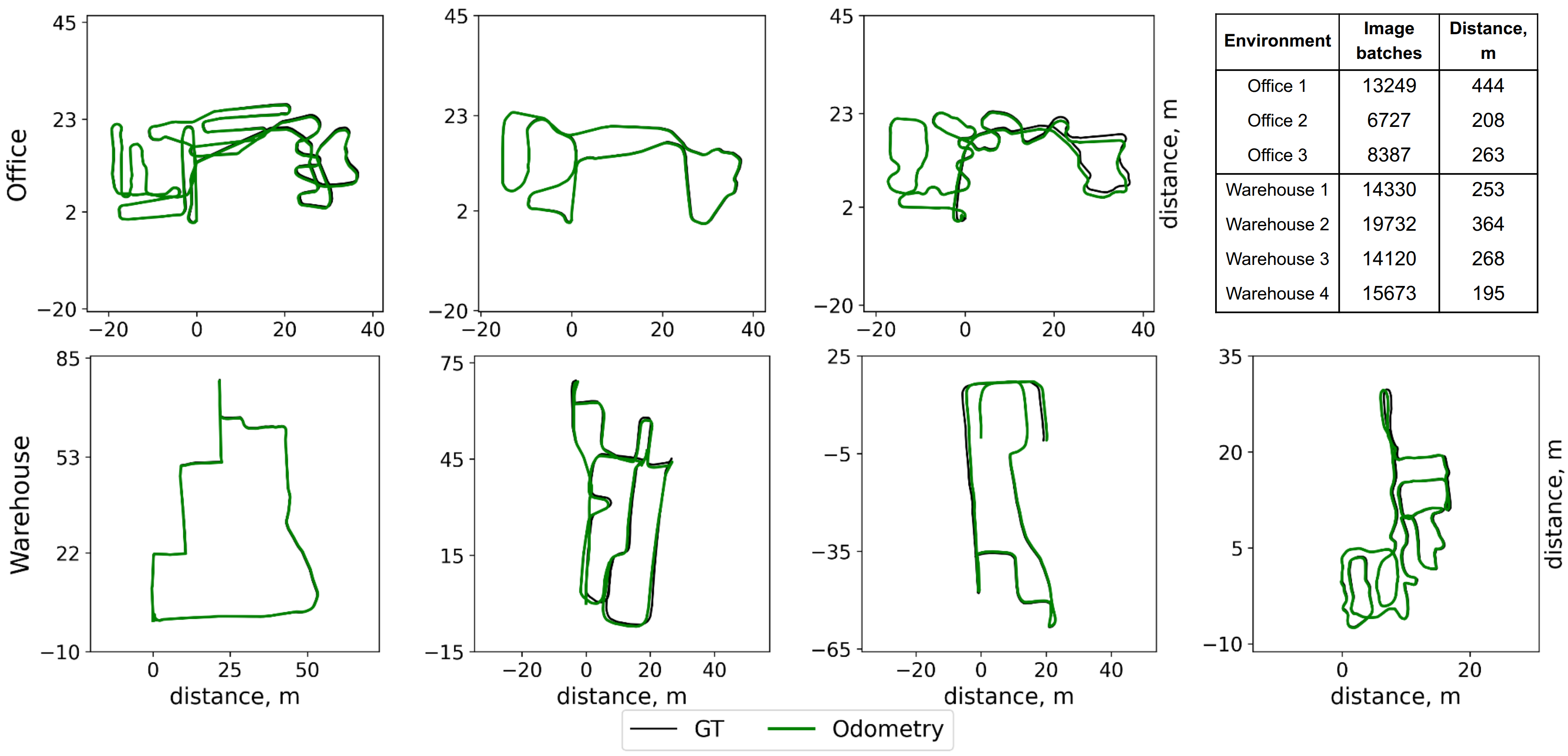}
    \caption{Trajectories predicted by cuVSLAM (4 Multi-Stereo Odometry) on the proprietary R2B wheeled robot dataset. Black lines represent ground truth; green lines represent odometry predictions. Attached table lists trajectory distances, number of 8-image batches, and trajectory index for environments (left to right).}
    \label{fig:r2b_trajectories}
\end{figure}

The Multi-Stereo R2B dataset was acquired using a wheeled Carter robot platform equipped with four stereo Hawk cameras, each capturing RGB images at a resolution of 1920×1200 pixels and 30 frames per second. The stereo camera pairs are hardware-synchronized to ensure temporal coherence, with synchronization precision maintained within 0.1 ms for each 8-image batch.
Data integrity validation confirmed minimal frame loss across all recordings, with the maximum batch loss rate remaining below 0.04\%. Figure~\ref{fig:r2b_trajectories} illustrates the complete set of recorded trajectories, along with corresponding trajectory lengths and synchronized image batch counts for each sequence.

\subsection{Appendix B: Hardware Scalability on Embedded Platforms} \label{sec:hardware_scalability}

This appendix presents an analysis of hardware utilization and visual tracking performance for cuVSLAM on embedded platforms, comparing Multi-Stereo and Mono-Depth modes as representative examples of sparse and dense feature-based approaches, respectively.

\textbf{Experimental Setup.}
The primary hardware configuration consisted of an NVIDIA Jetson Orin AGX operating in MAXN mode with 3 Intel RealSense D435 stereo cameras connected via USB and hardware-synchronized. Different resolutions were obtained directly from the cameras by modifying their configuration during the initialization step of each experiment. For Full HD resolution testing, we employed a set of 3 RGB Hawk stereo cameras connected via FAKRA2 connectors with hardware synchronization. All experiments were conducted in real-time with live camera feeds connected directly to the Jetson boards, and cuVSLAM calculations were performed on the same board in real-time. The system operated in Performance mode, utilizing the Isaac ROS (ROS2) stack for image acquisition and Multi-Stereo visual tracking, while the Python stack was employed for Mono-Depth visual tracking and processing.

\textbf{Methodology.}
Resource utilization was measured through a two-phase experimental protocol designed to isolate the computational overhead of cuVSLAM. In the first phase, system utilization was logged while receiving frames from freely moving cameras without invoking visual tracking, establishing a baseline measurement. In the second phase, the identical setup was used with cuVSLAM visual tracking enabled. CPU and GPU utilization metrics were sampled at 15ms intervals, and the time spent on cuVSLAM frame processing was logged for each frame and subsequently averaged. The net computational resource utilization of cuVSLAM was calculated by subtracting the baseline measurements from the active tracking measurements.

\begin{figure}[!h]
    \centering
	\captionsetup{justification=centering}
    \includegraphics[width=1.0\textwidth]{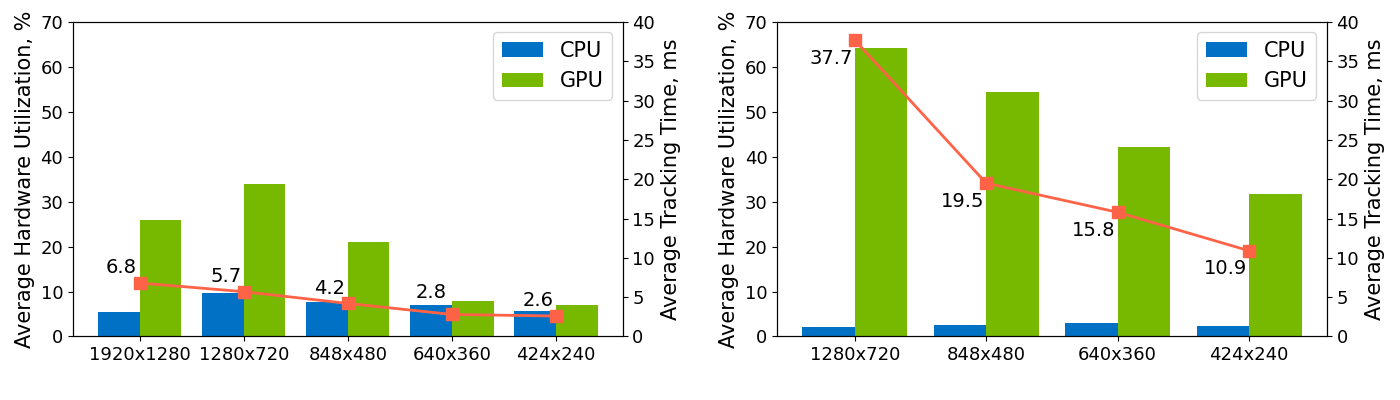}
    \caption{ cuVSLAM realtime network hardware utilization and callback time per input image on Jetson AGX Orin as a function of resolution for 3-Stereo Camera Visual Odometry (left) and Mono-Depth Visual Odometry (right) at 30 FPS. For further details, refer to Appendix ~\ref{sec:hardware_scalability}}
    \label{fig:hardware_utilization_AGX}
\end{figure}

\textbf{Jetson Orin AGX Performance.}
Figure~\ref{fig:hardware_utilization_AGX} illustrates the significant computational efficiency advantage of the sparse feature-based approach for tracking, demonstrating 3-7× faster processing times for 6-image batches compared to single image-depth pairs. The phenomenon where the 3-stereo camera visual tracking in Full HD configuration exhibited lower hardware utilization than HD configuration is attributed to differences in the camera field-of-view, with Hawk cameras provided 121.5° × 73.5° FOV in FHD mode versus RealSense D435 cameras provided 87° × 58° FOV in HD mode. The wider field-of-view affects the lifetime of keypoints as described in Section~\ref{sec:2d_module}, triggering more frequent keyframe creation and consequently invoking computationally intensive operations including 3D landmark triangulation and sparse bundle adjustment, as detailed in Sections~\ref{sec:2d_module} and ~\ref{sec:3d_module}.

For Mono-Depth mode, the average callback time for 720p resolution reached 37.7ms, which exceeds the 33.3ms frame interval required for 30 FPS operation. This timing constraint indicates GPU overload on the Orin AGX platform in this configuration, and according to our measurements, resulted in the system skipping approximately half of the incoming frames. Consequently, for HD resolution applications, we recommend either reducing the frame rate to 15 FPS or decreasing the resolution to VGA, as our experiments revealed no significant difference in cuVSLAM visual tracking quality between VGA and HD resolutions.

\textbf{Jetson Orin Nano Performance.}
To evaluate the feasibility of deploying cuVSLAM on entry-level hardware, we conducted experiments replacing the Jetson Orin AGX with a Jetson Orin Nano Super operating in 25W mode and utilizing exclusively the Python stack. The experiments explore possibility of using multi-stereo odometry mode for 640x360 resolution and 30 FPS followed the methodology described above with free camera movement.

\begin{figure}[!h]
    \centering
	\captionsetup{justification=centering}
    \includegraphics[width=0.6\textwidth]{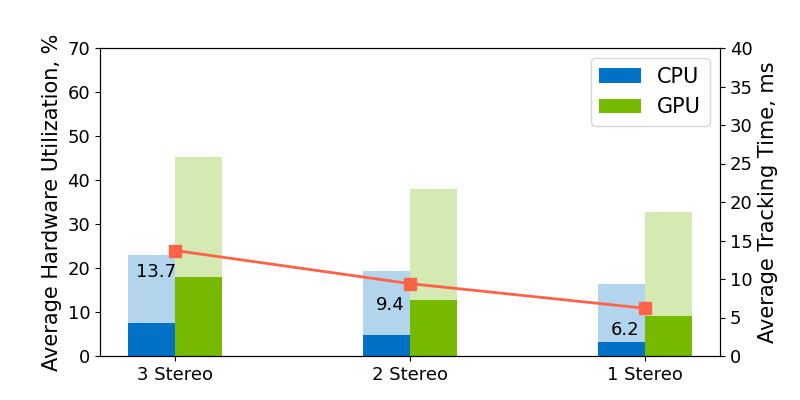}
    \caption{Hardware utilization and callback time on Jetson Orin Nano for Multi-Stereo Visual Odometry using RealSense cameras at 640×360 resolution and 30 FPS. Results are presented for varying numbers of cameras, showing total utilization (faded color bars) and net cuVSLAM utilization (normal color bars).}
    \label{fig:hardware_utilization_Nano}
\end{figure}

The results of average net cuVSLAM resource utilization, total system load, and average processing time are presented in Figure~\ref{fig:hardware_utilization_Nano}. The findings demonstrate substantial resource headroom, enabling the use of higher frame rates for configurations with one or 2 stereo cameras. For Mono-Depth processing at 30 FPS, stable operation was achieved only at the minimum resolution of 424×240 pixels. Although the average GPU load remained below 40\%, the observed average cuVSLAM call time of 31ms approached the critical 33ms frame interval at 30 FPS. When combined with additional operations in the image acquisition loop, this resulted in exceeding the frame interval and consequently dropping approximately 30\% of incoming frames.

It is important to note that for scenarios involving ground robots, the observed computational load will be lower than our free-movement test scenarios due to the reduced number of keyframes generated during predominantly inertial movements. However, for systems involving freely moving cameras with rapid scene changes, such as handheld cameras or unmanned aerial vehicles, we recommend increasing frame rates to 60 FPS when using RealSense cameras to ensure robust tracking performance. The selection of appropriate resolution and frame rate parameters should be carefully balanced based on the available computational resources and specific application requirements.

\begin{table}[h]
    \centering
    \caption{Comparative evaluation of odometry and SLAM performance for cuVSLAM Multi-Stereo (4 stereo cameras) on the TartanGround dataset. To assess the impact of SLAM mode, a subset of sequences containing looped trajectories with observable loop closures was selected. Reported values are averages computed per environment.}
    \label{tab:tartanairground_slam}
    \begin{tabular}{|l|l|l|c|c|c|}
    \hline
    Environment & Sequences & Mode & avgRTE & avgRE & RMSE APE \\
    \hline
    \multirow{2}{*}{\centering Downtown} & \multirow{2}{*}{\centering P7} & ODOM & 0.35 & 0.59 & 0.09 \\
     &  & SLAM & 0.25 & 0.41 & 0.09 \\
    \hline
    \multirow{2}{*}{\centering ForestEnv} & \multirow{2}{*}{\centering P1, P14, P17, P21} & ODOM & 0.42 & 0.78 & 0.16 \\
     &  & SLAM & 0.38 & 0.61 & 0.17 \\
    \hline
    \multirow{2}{*}{\centering Gascola} & \multirow{2}{*}{\centering P0, P3-P6, P9} & ODOM & 0.17 & 0.41 & 0.08 \\
     &  & SLAM & 0.13 & 0.31 & 0.06 \\
    \hline
    \multirow{2}{*}{\centering ModernCityDowntown} & \multirow{2}{*}{\centering \shortstack{P2, P5-P7}} & ODOM & 0.17 & 0.32 & 0.08 \\
     &  & SLAM & 0.12 & 0.24 & 0.06 \\
    \hline
    \multirow{2}{*}{\centering ModularNeighborhood} & \multirow{2}{*}{\centering \shortstack{P12, P20}} & ODOM & 0.42 & 1.17 & 0.14 \\
     &  & SLAM & 0.35 & 1.02 & 0.05 \\
    \hline
    \multirow{2}{*}{\centering NordicHarbor} & \multirow{2}{*}{\centering \shortstack{P1, P11, P17-P19\\P23, P27}} & ODOM & 0.17 & 0.41 & 0.08 \\
     &  & SLAM & 0.10 & 0.26 & 0.04 \\
    \hline
    \multirow{2}{*}{\centering OldTownFall} & \multirow{2}{*}{\centering P2} & ODOM & 0.07 & 0.31 & 0.05 \\
     &  & SLAM & 0.12 & 0.26 & 0.06 \\
    \hline
    \multirow{2}{*}{\centering OldTownSummer} & \multirow{2}{*}{\centering \shortstack{P2-P4}} & ODOM & 0.09 & 0.29 & 0.06 \\
     &  & SLAM & 0.06 & 0.25 & 0.03 \\
    \hline
    \multirow{2}{*}{\centering SeasonalForestAutumn} & \multirow{2}{*}{\centering \shortstack{P5, P8, P17}} & ODOM & 0.20 & 0.32 & 0.07 \\
     &  & SLAM & 0.19 & 0.33 & 0.07 \\
    \hline
    \multirow{2}{*}{\centering SeasonalForestSpring} & \multirow{2}{*}{\centering P9} & ODOM & 0.15 & 0.49 & 0.05 \\
     &  & SLAM & 0.11 & 0.39 & 0.02 \\
    \hline
    \multirow{2}{*}{\centering \textbf{Average}} & \multirow{2}{*}{\centering} & ODOM & \textbf{0.21} & \textbf{0.48} & \textbf{0.088} \\
     &  & SLAM & \textbf{0.17} & \textbf{0.37} & \textbf{0.065} \\
    \hline
    \end{tabular}
\end{table}

\begin{table}[h]
\centering
\begin{minipage}{0.48\textwidth}
\scriptsize
\caption{TartanAir V2 Hard cuVSLAM evaluation results (Envs 1--35, see Appendix \ref{sec:tartanairground_description} for details)}
\label{tab:tartanairv2_odom_1}
\begin{tabular}{|c|l|l|l|c|c|c|}
\hline
 & Env & Seq & Mode & RTE & RRE & APE \\
\hline
\multirow{2}{*}{1} & \multirow{2}{*}{\shortstack{Abandoned-\\Cable}} & \multirow{2}{*}{P0-P7} & ODOM & 1.90 & 13.77 & 8.97 \\
 &  &  & SLAM & 1.54 & 11.47 & 7.15 \\
\hline
\multirow{2}{*}{2} & \multirow{2}{*}{\shortstack{Abandoned-\\Factory}} & \multirow{2}{*}{\shortstack{P1-P6\\P8-P9}} & ODOM & 1.07 & 4.89 & 1.06 \\
 &  &  & SLAM & 0.77 & 3.70 & 0.95 \\
\hline
\multirow{2}{*}{3} & \multirow{2}{*}{\shortstack{Abandoned-\\Factory2}} & \multirow{2}{*}{P0-P5} & ODOM & 0.27 & 1.41 & 0.34 \\
 &  &  & SLAM & 0.19 & 0.84 & 0.24 \\
\hline
\multirow{2}{*}{4} & \multirow{2}{*}{\shortstack{Abandoned-\\School}} & \multirow{2}{*}{P0-P9} & ODOM & 3.59 & 33.78 & 5.41 \\
 &  &  & SLAM & 3.31 & 31.42 & 5.12 \\
\hline
\multirow{3}{*}{5} & \multirow{3}{*}{\shortstack{American-\\Diner}} & \multirow{3}{*}{\shortstack{P0, P2*\\P3-P4}} & ODOM* & 5.53 & 13.27 & 1.25 \\
 &  &  & ODOM & 5.21 & 11.12 & 1.05 \\
 &  &  & SLAM & 4.98 & 10.94 & 0.93 \\
\hline
\multirow{3}{*}{6} & \multirow{3}{*}{\shortstack{Amusement-\\Park}} & \multirow{3}{*}{\shortstack{P0, P1*\\P2-P3, P4*\\P5-P7}} & ODOM* & 1.15 & 3.74 & 1.53 \\
 &  &  & ODOM & 1.36 & 5.15 & 1.71 \\
 &  &  & SLAM & 1.36 & 5.49 & 1.71 \\
\hline
\multirow{2}{*}{7} & \multirow{2}{*}{\shortstack{Ancient-\\Towns}} & \multirow{2}{*}{P0-P6} & ODOM & 2.86 & 10.98 & 2.52 \\
 &  &  & SLAM & 2.50 & 8.93 & 2.23 \\
\hline
\multirow{3}{*}{8} & \multirow{3}{*}{\shortstack{Antiquity3D}} & \multirow{3}{*}{\shortstack{P1-P5, P6*\\P7, P8*,P9}} & ODOM* & 2.30 & 17.90 & 4.05 \\
 &  &  & ODOM & 1.95 & 16.53 & 3.69 \\
 &  &  & SLAM & 1.84 & 15.23 & 3.55 \\
\hline
\multirow{2}{*}{9} & \multirow{2}{*}{\shortstack{Apocalyptic}} & \multirow{2}{*}{P0-P4} & ODOM & 0.92 & 4.25 & 1.25 \\
 &  &  & SLAM & 0.66 & 2.81 & 0.98 \\
\hline
\multirow{3}{*}{10} & \multirow{3}{*}{\shortstack{ArchVizTiny-\\HouseDay}} & \multirow{3}{*}{\shortstack{P0, P1*\\P2-P3,P4*\\P5-P6}} & ODOM* & 3.75 & 4.02 & 0.37 \\
 &  &  & ODOM & 2.69 & 3.48 & 0.27 \\
 &  &  & SLAM & 2.44 & 3.14 & 0.25 \\
\hline
\multirow{3}{*}{11} & \multirow{3}{*}{\shortstack{ArchVizTiny-\\HouseNight}} & \multirow{3}{*}{\shortstack{P0*, P1\\P5, P6*\\P2-P4*}} & ODOM* & 5.37 & 6.23 & 0.47 \\
 &  &  & ODOM & 4.19 & 5.64 & 0.39 \\
 &  &  & SLAM & 4.15 & 5.91 & 0.36 \\
\hline
\multirow{2}{*}{12} & \multirow{2}{*}{\shortstack{Brushify-\\Moon}} & \multirow{2}{*}{P1-P5} & ODOM & 5.64 & 40.15 & 44.55 \\
 &  &  & SLAM & 6.23 & 41.28 & 45.50 \\
\hline
\multirow{2}{*}{13} & \multirow{2}{*}{\shortstack{CarWelding}} & \multirow{2}{*}{P0-P8} & ODOM & 1.14 & 5.74 & 1.34 \\
 &  &  & SLAM & 0.94 & 4.51 & 1.27 \\
\hline
\multirow{2}{*}{14} & \multirow{2}{*}{\shortstack{Castle-\\Fortress}} & \multirow{2}{*}{P0-P11} & ODOM & 0.93 & 7.33 & 2.65 \\
 &  &  & SLAM & 0.90 & 6.41 & 2.50 \\
\hline
\multirow{2}{*}{15} & \multirow{2}{*}{\shortstack{CoalMine}} & \multirow{2}{*}{P0-P5} & ODOM & 0.34 & 1.42 & 0.51 \\
 &  &  & SLAM & 0.20 & 0.67 & 0.38 \\
\hline
\multirow{3}{*}{16} & \multirow{3}{*}{\shortstack{Construct-\\Site}} & \multirow{3}{*}{\shortstack{P1, P3*\\P5*\\P6-P11}} & ODOM* & 2.80 & 8.81 & 3.15 \\
 &  &  & ODOM & 2.90 & 9.23 & 3.31 \\
 &  &  & SLAM & 2.80 & 8.74 & 3.14 \\
\hline
\multirow{2}{*}{17} & \multirow{2}{*}{\shortstack{Country-\\House}} & \multirow{2}{*}{P0-P5} & ODOM & 0.87 & 2.69 & 0.20 \\
 &  &  & SLAM & 0.16 & 0.62 & 0.05 \\
\hline
\multirow{2}{*}{18} & \multirow{2}{*}{\shortstack{CyberPunk-\\Downtown}} & \multirow{2}{*}{P0-P6} & ODOM & 1.19 & 7.27 & 4.01 \\
 &  &  & SLAM & 1.27 & 8.22 & 3.98 \\
\hline
\multirow{2}{*}{19} & \multirow{2}{*}{\shortstack{Cyberpunk}} & \multirow{2}{*}{P0-P7} & ODOM & 1.13 & 4.65 & 3.38 \\
 &  &  & SLAM & 1.13 & 5.43 & 4.23 \\
\hline
\multirow{2}{*}{20} & \multirow{2}{*}{\shortstack{Desert-\\GasStation}} & \multirow{2}{*}{P0-P4} & ODOM & 2.44 & 7.11 & 2.53 \\
 &  &  & SLAM & 2.31 & 6.78 & 2.43 \\
\hline
\multirow{2}{*}{21} & \multirow{2}{*}{\shortstack{Downtown}} & \multirow{2}{*}{P0-P8} & ODOM & 1.20 & 5.66 & 3.59 \\
 &  &  & SLAM & 1.21 & 5.52 & 3.35 \\
\hline
\multirow{3}{*}{22} & \multirow{3}{*}{\shortstack{Factory-\\Weather}} & \multirow{3}{*}{\shortstack{P1-P7, P8*\\P9-P12}} & ODOM* & 6.23 & 39.06 & 22.37 \\
 &  &  & ODOM & 6.31 & 40.09 & 22.64 \\
 &  &  & SLAM & 6.17 & 37.40 & 22.22 \\
\hline
\multirow{2}{*}{23} & \multirow{2}{*}{\shortstack{Fantasy}} & \multirow{2}{*}{P0-P6} & ODOM & 1.01 & 4.16 & 1.45 \\
 &  &  & SLAM & 1.01 & 4.70 & 1.87 \\
\hline
\multirow{3}{*}{24} & \multirow{3}{*}{\shortstack{ForestEnv}} & \multirow{3}{*}{\shortstack{P0, P2-P5\\P7, P9\\P10*, P11}} & ODOM* & 3.67 & 21.93 & 13.76 \\
 &  &  & ODOM & 3.55 & 21.75 & 14.42 \\
 &  &  & SLAM & 3.62 & 23.64 & 15.47 \\
\hline
\multirow{2}{*}{25} & \multirow{2}{*}{\shortstack{Gascola}} & \multirow{2}{*}{P0-P9} & ODOM & 2.36 & 14.83 & 5.47 \\
 &  &  & SLAM & 2.49 & 14.56 & 6.13 \\
\hline
\multirow{2}{*}{26} & \multirow{2}{*}{\shortstack{Gothic-\\Island}} & \multirow{2}{*}{P0-P10} & ODOM & 1.68 & 13.19 & 4.00 \\
 &  &  & SLAM & 1.44 & 11.60 & 3.54 \\
\hline
\multirow{2}{*}{27} & \multirow{2}{*}{\shortstack{HQWestern-\\Saloon}} & \multirow{2}{*}{P0-P7} & ODOM & 2.67 & 16.93 & 2.01 \\
 &  &  & SLAM & 1.96 & 8.81 & 1.60 \\
\hline
\multirow{3}{*}{28} & \multirow{3}{*}{\shortstack{HongKong}} & \multirow{3}{*}{\shortstack{P0, P1*\\P2-P4}} & ODOM* & 5.26 & 18.85 & 5.02 \\
 &  &  & ODOM & 4.81 & 18.85 & 3.99 \\
 &  &  & SLAM & 4.75 & 17.78 & 3.89 \\
\hline
\multirow{2}{*}{29} & \multirow{2}{*}{\shortstack{Hospital}} & \multirow{2}{*}{P0-P10} & ODOM & 1.50 & 16.17 & 3.25 \\
 &  &  & SLAM & 1.30 & 14.85 & 2.93 \\
\hline
\multirow{2}{*}{30} & \multirow{2}{*}{\shortstack{House}} & \multirow{2}{*}{P0-P7} & ODOM & 1.47 & 9.80 & 0.47 \\
 &  &  & SLAM & 0.98 & 6.43 & 0.26 \\
\hline
\multirow{2}{*}{31} & \multirow{2}{*}{\shortstack{Industrial-\\Hangar}} & \multirow{3}{*}{P0-P6} & ODOM & 3.43 & 19.61 & 8.21 \\
 &  &  & SLAM & 3.48 & 19.86 & 8.03 \\
\hline
\multirow{2}{*}{32} & \multirow{2}{*}{\shortstack{Japanese-\\Alley}} & \multirow{2}{*}{P0-P6} & ODOM & 0.94 & 5.27 & 0.48 \\
 &  &  & SLAM & 0.34 & 1.87 & 0.39 \\
\hline
\multirow{2}{*}{33} & \multirow{2}{*}{\shortstack{Japanese-\\City}} & \multirow{2}{*}{P0-P10} & ODOM & 1.81 & 7.89 & 2.08 \\
 &  &  & SLAM & 1.74 & 7.80 & 2.03 \\
\hline
\multirow{2}{*}{34} & \multirow{2}{*}{\shortstack{MiddleEast}} & \multirow{2}{*}{\shortstack{P0-P10}} & ODOM & 2.96 & 13.18 & 2.64 \\
 &  &  & SLAM & 2.65 & 11.34 & 2.47 \\
\hline
\multirow{2}{*}{35} & \multirow{2}{*}{\shortstack{Mod-\\UrbanCity}} & \multirow{2}{*}{P0-P11} & ODOM & 3.65 & 23.51 & 5.74 \\
 &  &  & SLAM & 3.70 & 23.22 & 5.60 \\
\hline
\end{tabular}
\end{minipage}%
\hfill
\begin{minipage}{0.48\textwidth}
\scriptsize
\caption{TartanAir V2 Hard cuVSLAM evaluation results (Envs 36--71, see Appendix \ref{sec:tartanairground_description} for details)}
\label{tab:tartanairv2_odom_2}
\begin{tabular}{|c|l|l|l|c|c|c|}
\hline
 & Env & Seq & Mode & RTE & RRE & APE \\
\hline

\multirow{2}{*}{36} & \multirow{2}{*}{\shortstack{ModernCity\\Downtown}} & \multirow{2}{*}{P0-P8} & ODOM & 2.34 & 11.70 & 4.01 \\
 &  &  & SLAM & 2.08 & 11.34 & 3.60 \\
 \hline
 \multirow{2}{*}{37} & \multirow{2}{*}{\shortstack{Modular-\\Neighborhood}} & \multirow{2}{*}{P0-P10} & ODOM & 4.14 & 22.48 & 17.10 \\
 &  &  & SLAM & 3.60 & 21.04 & 13.85 \\
\hline
\multirow{2}{*}{38} & \multirow{2}{*}{\shortstack{ModularNeigh\\borhoodIntExt}} & \multirow{2}{*}{\shortstack{P1-P5\\P7-P8}} & ODOM & 2.64 & 15.80 & 4.74 \\
 &  &  & SLAM & 2.62 & 13.93 & 5.13 \\
\hline
\multirow{2}{*}{39} & \multirow{2}{*}{\shortstack{NordicHarbor}} & \multirow{2}{*}{P0-P7} & ODOM & 2.07 & 14.37 & 6.04 \\
 &  &  & SLAM & 1.97 & 15.04 & 5.77 \\
\hline
\multirow{3}{*}{40} & \multirow{3}{*}{\shortstack{Ocean}} & \multirow{3}{*}{\shortstack{P1-P3\\P4*\\P5-P7}} & ODOM* & 1.76 & 8.13 & 1.99 \\
 &  &  & ODOM & 1.65 & 8.66 & 2.36 \\
 &  &  & SLAM & 1.47 & 7.12 & 2.22 \\
\hline
\multirow{2}{*}{41} & \multirow{2}{*}{\shortstack{Office}} & \multirow{2}{*}{P0-P6} & ODOM* & 3.51 & 16.02 & 2.71 \\
 &  &  & SLAM & 3.36 & 14.84 & 2.49 \\
\hline
\multirow{3}{*}{42} & \multirow{3}{*}{\shortstack{OldBrick-\\HouseDay}} & \multirow{3}{*}{\shortstack{P1-P2\\P3*\\P4-P7}} & ODOM* & 1.74 & 11.15 & 1.48 \\
 &  &  & ODOM & 1.54 & 11.80 & 1.25 \\
 &  &  & SLAM & 1.29 & 7.76 & 1.10 \\
\hline
\multirow{3}{*}{43} & \multirow{3}{*}{\shortstack{OldBrick-\\HouseNight}} & \multirow{3}{*}{\shortstack{P1*, P2\\P4-P7}} & ODOM* & 1.83 & 16.07 & 1.53 \\
 &  &  & ODOM & 1.68 & 16.35 & 1.56 \\
 &  &  & SLAM & 1.59 & 13.05 & 1.50 \\
\hline
\multirow{2}{*}{44} & \multirow{2}{*}{\shortstack{Old-\\IndustrialCity}} & \multirow{2}{*}{P0-P9} & ODOM & 2.25 & 8.60 & 4.33 \\
 &  &  & SLAM & 1.58 & 7.20 & 3.76 \\
\hline
\multirow{2}{*}{45} & \multirow{2}{*}{\shortstack{Old-\\Scandinavia}} & \multirow{2}{*}{P1-P6} & ODOM* & 2.22 & 11.68 & 8.90 \\
 &  &  & SLAM & 2.15 & 11.47 & 8.38 \\
\hline
\multirow{3}{*}{46} & \multirow{3}{*}{\shortstack{OldTownFall}} & \multirow{3}{*}{\shortstack{P0*,P1*\\P2}} & ODOM* & 1.62 & 6.44 & 1.53 \\
 &  &  & ODOM & 1.25 & 4.09 & 0.11 \\
 &  &  & SLAM & 0.75 & 2.47 & 0.28 \\
\hline
\multirow{2}{*}{47} & \multirow{2}{*}{\shortstack{OldTownNight}} & \multirow{2}{*}{P0-P2} & ODOM & 0.83 & 4.74 & 1.24 \\
 &  &  & SLAM & 0.61 & 3.49 & 0.95 \\
\hline
\multirow{2}{*}{48} & \multirow{2}{*}{\shortstack{OldTown-\\Summer}} & \multirow{2}{*}{P0-P2} & ODOM & 2.42 & 11.90 & 1.57 \\
 &  &  & SLAM & 2.41 & 11.84 & 1.54 \\
\hline
\multirow{2}{*}{49} & \multirow{2}{*}{\shortstack{OldTown-\\Winter}} & \multirow{2}{*}{P0-P2} & ODOM & 0.60 & 2.14 & 0.72 \\
 &  &  & SLAM & 0.53 & 2.33 & 0.75 \\
\hline
\multirow{2}{*}{50} & \multirow{2}{*}{\shortstack{PolarSciFi}} & \multirow{2}{*}{P0-P8} & ODOM & 2.76 & 12.74 & 2.19 \\
 &  &  & SLAM & 2.68 & 12.34 & 2.18 \\
\hline
\multirow{2}{*}{51} & \multirow{2}{*}{\shortstack{Prison}} & \multirow{2}{*}{P0-P11} & ODOM & 4.14 & 21.37 & 4.59 \\
 &  &  & SLAM & 4.12 & 20.78 & 4.68 \\
\hline
\multirow{2}{*}{52} & \multirow{2}{*}{\shortstack{Restaurant}} & \multirow{2}{*}{P1-P8} & ODOM & 1.51 & 10.99 & 0.97 \\
 &  &  & SLAM & 1.03 & 8.00 & 0.70 \\
\hline
\multirow{2}{*}{53} & \multirow{2}{*}{\shortstack{RetroOffice}} & \multirow{2}{*}{P0-P5} & ODOM & 1.14 & 3.11 & 0.10 \\
 &  &  & SLAM & 0.94 & 2.85 & 0.09 \\
\hline
\multirow{2}{*}{54} & \multirow{2}{*}{\shortstack{Ruins}} & \multirow{2}{*}{\shortstack{P1-P6\\P8-P9}} & ODOM & 3.07 & 16.68 & 9.43 \\
 &  &  & SLAM & 3.05 & 17.07 & 9.09 \\
\hline
\multirow{2}{*}{55} & \multirow{2}{*}{\shortstack{SeasideTown}} & \multirow{2}{*}{P0-P7} & ODOM & 0.67 & 1.79 & 0.56 \\
 &  &  & SLAM & 0.61 & 1.58 & 0.44 \\
\hline
\multirow{2}{*}{56} & \multirow{2}{*}{\shortstack{SeasonForest-\\Autumn}} & \multirow{2}{*}{P0-P2} & ODOM & 5.41 & 35.20 & 21.38 \\
 &  &  & SLAM & 5.10 & 31.76 & 18.73 \\
\hline
\multirow{2}{*}{57} & \multirow{2}{*}{\shortstack{SeasonForest-\\Spring}} & \multirow{2}{*}{P0-P3} & ODOM & 4.09 & 25.14 & 15.03 \\
 &  &  & SLAM & 3.93 & 28.30 & 15.02 \\
\hline
\multirow{2}{*}{58} & \multirow{2}{*}{\shortstack{SeasonForest-\\SummerNight}} & \multirow{2}{*}{P0-P1} & ODOM & 3.65 & 21.45 & 25.05 \\
 &  &  & SLAM & 3.45 & 19.42 & 23.00 \\
\hline
\multirow{2}{*}{59} & \multirow{2}{*}{\shortstack{SeasonForest-\\Winter}} & \multirow{2}{*}{P0-P2} & ODOM & 6.73 & 30.59 & 15.47 \\
 &  &  & SLAM & 7.53 & 32.61 & 19.11 \\
\hline
\multirow{2}{*}{60} & \multirow{2}{*}{\shortstack{SeasonForest-\\WinterNight}} & \multirow{2}{*}{P0-P2} & ODOM & 4.77 & 30.45 & 27.40 \\
 &  &  & SLAM & 5.14 & 30.09 & 26.31 \\
\hline
\multirow{2}{*}{61} & \multirow{2}{*}{\shortstack{Sewerage}} & \multirow{2}{*}{P1-P11} & ODOM & 3.94 & 30.11 & 6.74 \\
 &  &  & SLAM & 3.87 & 27.64 & 5.92 \\
\hline
\multirow{2}{*}{62} & \multirow{2}{*}{\shortstack{ShoreCaves}} & \multirow{2}{*}{P1-P8} & ODOM & 1.86 & 7.56 & 3.41 \\
 &  &  & SLAM & 1.77 & 7.20 & 3.00 \\
\hline
\multirow{3}{*}{63} & \multirow{3}{*}{\shortstack{Slaughter}} & \multirow{3}{*}{\shortstack{P0, P1*\\P2-P10\\P11*}} & ODOM* & 4.52 & 26.89 & 6.04 \\
 &  &  & ODOM & 3.97 & 27.19 & 5.53 \\
 &  &  & SLAM & 3.80 & 24.49 & 5.34 \\
\hline
\multirow{3}{*}{64} & \multirow{3}{*}{\shortstack{SoulCity}} & \multirow{3}{*}{\shortstack{P0*,P1,P2*\\P3-P4, P5*\\P6, P7}} & ODOM* & 5.22 & 32.59 & 7.30 \\
 &  &  & ODOM & 4.74 & 28.79 & 7.96 \\
 &  &  & SLAM & 4.82 & 29.27 & 8.10 \\
\hline
\multirow{3}{*}{65} & \multirow{3}{*}{\shortstack{Supermarket}} & \multirow{3}{*}{\shortstack{P0*,P1,P2\\P3*, P4-P6}} & ODOM* & 4.70 & 23.19 & 3.69 \\
 &  &  & ODOM & 4.82 & 22.64 & 3.66 \\
 &  &  & SLAM & 4.73 & 23.38 & 3.60 \\
\hline
\multirow{2}{*}{66} & \multirow{2}{*}{\shortstack{Terrain-\\Blending}} & \multirow{2}{*}{\shortstack{P1, P3-P4}} & ODOM & 0.45 & 1.65 & 1.31 \\
 &  &  & SLAM & 0.55 & 1.90 & 1.55 \\
\hline
\multirow{2}{*}{67} & \multirow{2}{*}{\shortstack{Urban-\\Construction}} & \multirow{2}{*}{P0-P4} & ODOM & 0.35 & 1.76 & 1.30 \\
 &  &  & SLAM & 0.31 & 1.50 & 1.24 \\
\hline
\multirow{2}{*}{68} & \multirow{2}{*}{\shortstack{VictorianStreet}} & \multirow{2}{*}{P0-P7} & ODOM & 0.99 & 4.67 & 0.43 \\
 &  &  & SLAM & 0.72 & 3.17 & 0.25 \\
\hline
\multirow{2}{*}{69} & \multirow{2}{*}{\shortstack{WaterMillDay}} & \multirow{2}{*}{\shortstack{P1-P3, P6}} & ODOM & 0.48 & 2.54 & 0.74 \\
 &  &  & SLAM & 0.30 & 1.12 & 0.59 \\
\hline
\multirow{2}{*}{70} & \multirow{2}{*}{\shortstack{WaterMillNight}} & \multirow{2}{*}{P0-P6} & ODOM & 0.34 & 0.94 & 0.78 \\
 &  &  & SLAM & 0.29 & 0.92 & 0.72 \\
\hline
\multirow{2}{*}{71} & \multirow{2}{*}{\shortstack{Western-\\DesertTown}} & \multirow{2}{*}{\shortstack{P1-P7, P9\\P11, P13}} & ODOM & 1.89 & 13.77 & 8.09 \\
 &  &  & SLAM & 1.78 & 12.97 & 7.32 \\
\hline
\end{tabular}
\end{minipage}
\end{table}


\end{document}

%% file: packages.tex
\usepackage[authoryear,sort&compress,round]{natbib}

\usepackage[utf8]{inputenc} 
\usepackage[T1]{fontenc}    

\usepackage{parskip}        
\usepackage{url}            
\usepackage{booktabs}       
\usepackage{amsfonts}       
\usepackage{nicefrac}       
\usepackage{microtype}      
\usepackage{xcolor}         
\usepackage[dvipsnames]{xcolor} 
\usepackage{graphicx}
\usepackage{animate}        
\usepackage{subcaption}
\usepackage{tabularx}
\usepackage{makecell}
\usepackage{adjustbox}
\usepackage{setspace}
\newcolumntype{M}[1]{>{\centering\arraybackslash}m{#1}}
\usepackage{float}
\usepackage{tikz}
\usetikzlibrary{positioning,shapes,arrows}
\usepackage{amsmath,amsfonts,bm, bbm,leftindex}
\usepackage{multirow}
\usepackage{comment}
\usepackage{gensymb}
\usepackage{lipsum}
\usetikzlibrary{arrows.meta, positioning, fit}
\usepackage[para]{threeparttable}
\usepackage{tikz}
\usetikzlibrary{tikzmark}



%% file: common.tex
\def\eqref#1{equation~\ref{#1}}

\def\1{\bm{1}}

\DeclareMathAlphabet{\mathsfit}{\encodingdefault}{\sfdefault}{m}{sl}
\SetMathAlphabet{\mathsfit}{bold}{\encodingdefault}{\sfdefault}{bx}{n}

\makeatletter
\let\save@mathaccent\mathaccent
\newcommand*\if@single[3]{%
  \setbox0\hbox{${\mathaccent"0362{#1}}^H$}%
  \setbox2\hbox{${\mathaccent"0362{\kern0pt#1}}^H$}%
  \ifdim\ht0=\ht2 #3\else #2\fi
  }
\newcommand*\rel@kern[1]{\kern#1\dimexpr\macc@kerna}
\newcommand*\widebar[1]{\@ifnextchar^{{\wide@bar{#1}{0}}}{\wide@bar{#1}{1}}}
\newcommand*\wide@bar[2]{\if@single{#1}{\wide@bar@{#1}{#2}{1}}{\wide@bar@{#1}{#2}{2}}}
\newcommand*\wide@bar@[3]{%
  \begingroup
  \def\mathaccent##1##2{%
    \let\mathaccent\save@mathaccent
    \if#32 \let\macc@nucleus\first@char \fi
    \setbox\z@\hbox{$\macc@style{\macc@nucleus}_{}$}%
    \setbox\tw@\hbox{$\macc@style{\macc@nucleus}{}_{}$}%
    \dimen@\wd\tw@
    \advance\dimen@-\wd\z@
    \divide\dimen@ 3
    \@tempdima\wd\tw@
    \advance\@tempdima-\scriptspace
    \divide\@tempdima 10
    \advance\dimen@-\@tempdima
    \ifdim\dimen@>\z@ \dimen@0pt\fi
    \rel@kern{0.6}\kern-\dimen@
    \if#31
      \overline{\rel@kern{-0.6}\kern\dimen@\macc@nucleus\rel@kern{0.4}\kern\dimen@}%
      \advance\dimen@0.4\dimexpr\macc@kerna
      \let\final@kern#2%
      \ifdim\dimen@<\z@ \let\final@kern1\fi
      \if\final@kern1 \kern-\dimen@\fi
    \else
      \overline{\rel@kern{-0.6}\kern\dimen@#1}%
    \fi
  }%
  \macc@depth\@ne
  \let\math@bgroup\@empty \let\math@egroup\macc@set@skewchar
  \mathsurround\z@ \frozen@everymath{\mathgroup\macc@group\relax}%
  \macc@set@skewchar\relax
  \let\mathaccentV\macc@nested@a
  \if#31
    \macc@nested@a\relax111{#1}%
  \else
    \def\gobble@till@marker##1\endmarker{}%
    \futurelet\first@char\gobble@till@marker#1\endmarker
    \ifcat\noexpand\first@char A\else
      \def\first@char{}%
    \fi
    \macc@nested@a\relax111{\first@char}%
  \fi
  \endgroup
}
\makeatother
